\documentclass[runningheads]{llncs}

\usepackage{eccv}

\usepackage{eccvabbrv}

\usepackage{graphicx}
\usepackage{booktabs}

\usepackage{algorithm}
\usepackage{booktabs}
\usepackage{newfloat}
\usepackage{listings}
\usepackage{multirow}
\usepackage{amsmath}
\usepackage{booktabs}
\usepackage{amsfonts}
\usepackage{xcolor,colortbl}

\usepackage{array}     
\usepackage{makecell}
\usepackage{pifont}
\usepackage{adjustbox}

\usepackage{graphicx}
\usepackage{booktabs}

\usepackage{algpseudocode}
\usepackage{rotating}
\usepackage{siunitx}

\usepackage[accsupp]{axessibility}  

\usepackage{hyperref}

\usepackage{orcidlink}

\begin{document}

\title{FeDepth: Federated Learning for Depth Estimation under Robot Heterogeneity} 

\newcommand{\before}[1]{{\color{red}#1}}
\newcommand{\gang}[1]{{\textcolor{blue}{Gang$>$#1}}}
\newcommand{\inha}[1]{{\textcolor{purple}{Inha$>$#1}}}
\newcommand{\eon}[1]{{\textcolor{green}{Jeongeon$>$#1}}}
\newcommand{\jun}[1]{{\textcolor{magenta}{Junhee$>$#1}}}
\newcommand{\KDJ}[1]{\textcolor{red}{Joo$>$#1}}

\titlerunning{FeDepth}


\author{Ganghyeon Lee\inst{*}\orcidlink{0009-0006-7046-4072} \and
Inha Lee\inst{*}\orcidlink{0009-0003-8030-3131} \and
Junhee Lee\inst{}\orcidlink{0009-0001-4078-8811} \and
Jeongeon Lee\inst{}\orcidlink{0009-0007-7474-2620} \and \\
Sungwhan Yoon\inst{}\orcidlink{0000-0002-7202-2837}  \and
Kyungdon Joo\inst{\dagger}\orcidlink{0000-0002-3920-9608}}

\authorrunning{G. Lee \textit{et al.}}

\institute{Ulsan National Institute of Science and Technology (UNIST), Republic of Korea \\
    \email{\{create0327,epsilon8854,junhee98,aeonian,shyoon8,kyungdon\}@unist.ac.kr}}
\maketitle
\begingroup
\makeatletter
\renewcommand{\@makefnmark}{\hbox{\textsuperscript{*}}}
\makeatother
\footnotetext[1]{Equal contribution. \quad
\textsuperscript{\textdagger} Corresponding author.}
\endgroup
\vspace{-4mm} 
\begin{abstract}
Although recent robot perception research emphasizes training on data from diverse environments to improve generalization, most existing methods still rely on centralized learning, which is inefficient and difficult to scale across heterogeneous robot platforms. Federated learning (FL) offers an alternative by enabling distributed training without raw data transfer, but it suffers from severe performance degradation under domain shifts caused by heterogeneity across clients.
In real robotic deployments, data distributions often overlap across platforms, environments, and sensing conditions, making it difficult to partition clients into clearly separated domains. 
However, this characteristic breaks the assumption of clearly separable client domains commonly used in clustered FL.
%
To address this gap in robot perception, particularly in depth estimation, we introduce two realistic and unexplored non-IID scenarios that reflect heterogeneity in terms of platform, environment, and depth distribution. 
We then propose \textsc{FeDepth}, a descriptor-based clustered FL framework that models client relationships through soft clustering.
Unlike hard clustering methods that assume clearly separated clusters, 
\textsc{FeDepth} allows clients to participate in multiple clusters, capturing continuous and ambiguous domain transitions commonly observed in robotic environments.
%
Extensive experiments demonstrate that \textsc{FeDepth} consistently improves robustness over standard FL and clustered FL baselines across multiple depth estimation architectures, providing a practical and effective solution for federated robot perception.
Our project page is available at \url{https://vision3d-lab.github.io/fedepth/}.
%
%
%

\vspace{-2mm} 
  \keywords{Federated Learning \and Robot Perception \and Depth Estimation}
\end{abstract}
   
\begin{figure}[!t]
    \centering
    \includegraphics[width=0.9\linewidth]{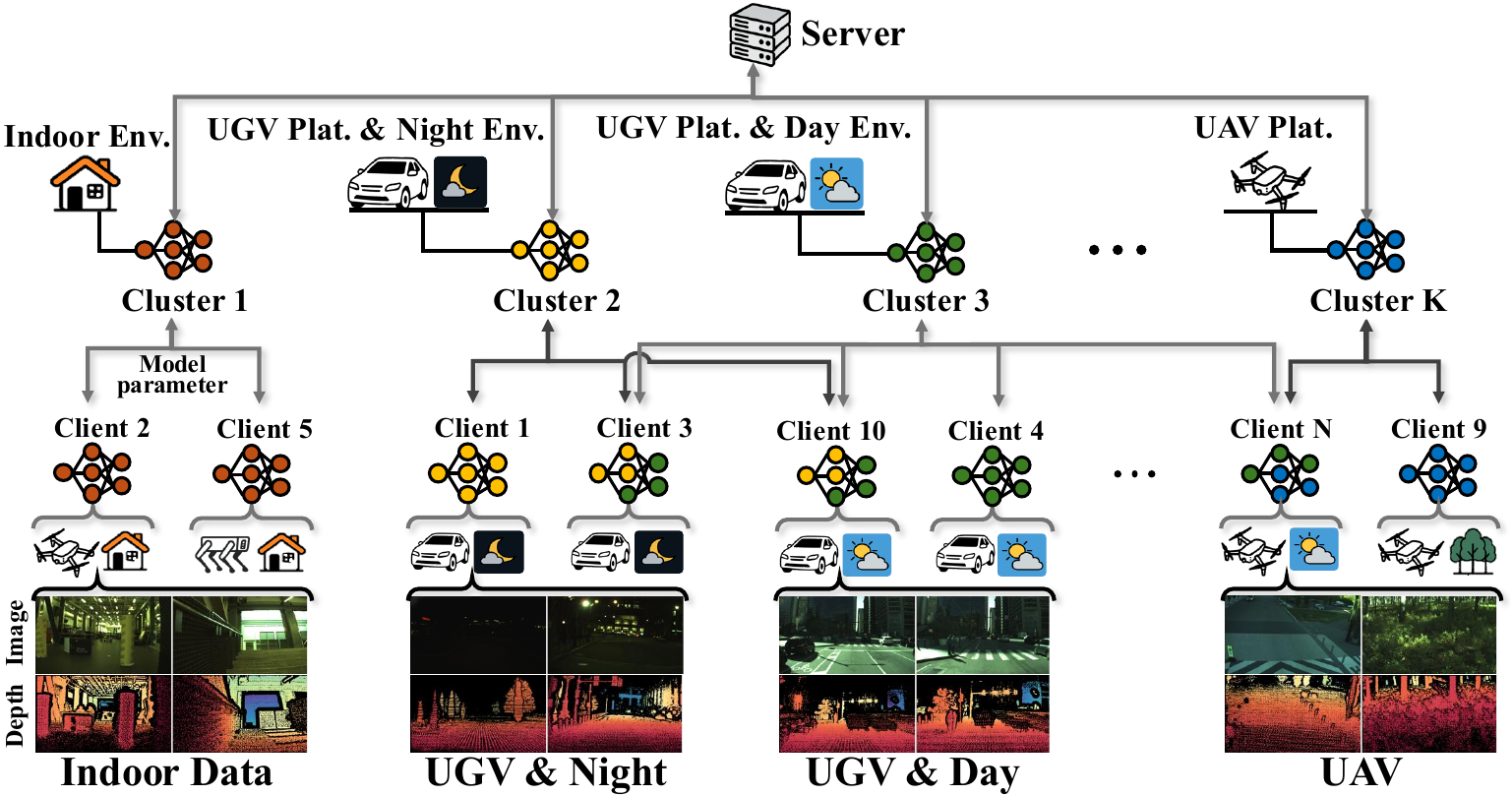}
    \vspace{-1mm} 
    \caption{{\textbf{Illustration of \textsc{FeDepth}.}
    The proposed \textsc{FeDepth}, a soft clustering-based FL framework that models complex inter-client relationships for robust training and effective cluster model updates.
    For example, Clients 3 and 10, which share the same platform (\emph{e.g.}, UGV), are assigned to two overlapping clusters.}
    %
    %
    %
    }
    \vspace{-4mm} 
    \label{fig:teasure}
\end{figure} 

\vspace{-4mm} 
\section{Introduction}
\vspace{-2mm} 
Robot perception, including tasks such as depth estimation~\cite{adabins}, object recognition~\cite{ren2016faster}, and semantic understanding~\cite{chen2018encoder}, is fundamental to enabling autonomous systems to interact safely and intelligently with their surroundings. Recent advances in deep learning, particularly convolutional neural networks (CNNs) and Transformer-based architectures, have notably improved perception performance across various modalities~\cite{gehrig2023recurrent, zhang2023cmx, misra2021end}. 
In addition, training on datasets collected from diverse environments~\cite{kirillov2023segment, yang2024depth} has been shown to enhance generalization beyond controlled laboratory conditions to the complex and variable settings encountered in real-world robotic applications.

Despite these advances, most perception pipelines still rely on a centralized learning (CL) paradigm: (\emph{i}) collecting raw sensory data in the field, (\emph{ii}) transmitting it to a central server, and (\emph{iii}) training a single model on the aggregated dataset. 
Although this approach has proven effective in controlled benchmarks with curated data, it becomes impractical when training on data acquired from real-world robots.
%
Concretely, CL suffers from several limitations, including high communication overhead incurred by transmitting high-bandwidth sensory data (\emph{e.g.}, high-resolution images or LiDAR streams) to a central server. 
Furthermore, CL faces severe scalability bottlenecks, as the computational requirements of servers scale proportionally with the volume of the aggregated dataset.
Moreover, CL is highly vulnerable to increasing privacy concerns when collecting visual perception data from personal robotic agents.

Federated learning (FL)~\cite{fedavg} provides a promising alternative: each robot trains its model locally and shares only weights rather than raw images. 
Although FL is widely known to perform well for canonical image classification benchmarks  (\emph{e.g.}, CIFAR-100~\cite{cifar100}), our preliminary experiments reveal that FL in robot perception scenarios inherently suffers from severe heterogeneity, which is driven by the diversity of robot platforms~(\emph{e.g.}, UGV, UAV, legged robots), various observation environments (\emph{e.g.}, urban, forest, indoor), and unbalanced data quantities across clients, thereby raising a significant challenge to deploying FL in real-world robots.
It should be noted that only a few recent works have pushed FL to robot perception applications ~\cite{kou2025fedema,liu2020fedvision, poggi2024federated}.
However, these studies are confined to a single platform (\emph{e.g.}, manipulator) or a limited environment (\emph{e.g.}, on-road driving), thus failing to address the substantial domain gaps induced by heterogeneous real-world robots.
Moreover, data distributions across robots are often partially overlapping rather than clearly separable, as robots may share similar environments, sensing conditions, or platform characteristics, suggesting the need for soft clustering that allows overlapping client assignments.

In this work, we focus on two real-world scenarios that encompass the domain shifts frequently encountered in heterogeneous robot data.
First, a heterogeneous platform-environment (HPE) scenario incorporates the heterogeneity originating from a unique combination of platform and environment for each client's trajectory.
The combinatorial configurations induce variations in viewpoint, motion dynamics, and scene layout, thus leading to significant domain gaps that hinder federated optimization.
Second, a bi-modal range (BMR) scenario contains a bi-modal distribution over depth ranges, distinctly separated into indoor and outdoor domains. Indoor clients observe close-range geometry (under 10 meters), while outdoor clients capture long-range scenes (up to 80 meters), creating a fundamental mismatch in depth distribution. This divergence, coupled with differing camera intrinsics, severely limits the performance of unified models even under centralized training.



To address these heterogeneities, we propose $\textsc{FeDepth}$, a clustered-FL (CFL) framework for monocular depth estimation on heterogeneous multi-robot systems (see Fig.~\ref{fig:teasure}). 
%
%
%
In $\textsc{FeDepth}$, we effectively tackle heterogeneous data by grouping clients based on their data distribution and performing intra-group aggregation.
This approach ensures that each cluster consists of clients with relatively homogeneous data, which facilitates robust federated optimization even under heterogeneity.
This principle is rooted in CFL~\cite{sattler2020clustered}, and we observe that a CFL paradigm can mitigate the challenges of robot-driven heterogeneity.
%
However, these conventional CFL approaches do not yield stable performance gains across all baselines.
Furthermore, clients in real-world scenarios often share multiple attributes, making it suboptimal or even infeasible to assign a client to a single, unique group.
In contrast, $\textsc{FeDepth}$ permits a client to be associated with multiple groups, enabling a single client to contribute to multiple relevant clusters.
To this end, we also propose a client clustering algorithm capable of performing this multi-cluster assignment.
We apply $\textsc{FeDepth}$ to depth estimation, which is a fundamental robot perception task that spans diverse architectures and representations (\emph{e.g.}, bins~\cite{adabins}, frequency~\cite{dcdepth}).
%
We evaluate $\textsc{FeDepth}$ on the HPE and BMR scenarios, where 
$\textsc{FeDepth}$ generalizes across datasets as well as depth estimation models. Our main contributions are:
\vspace{-1mm}
\begin{itemize}
    \item {We systematically investigate the impact of robot-driven heterogeneity on FL, a critical but under-explored challenge in robot perception.}
    \item We design two novel non-IID scenarios (HPE and BMR) derived from robot perception datasets. These scenarios establish a new realistic testbed for benchmarking federated depth estimation under robot-driven heterogeneity. 
    \item We propose \textsc{FeDepth}, a CFL framework that resolves robot-driven heterogeneity using soft cluster assignments. 
    We can handle overlapping client distributions 
    that conventional CFL methods cannot effectively manage.
    \item We validate that \textsc{FeDepth} is model-agnostic, demonstrating 
    generalization by integrating it with diverse depth estimation networks and representations. 
\end{itemize}

\section{Related Work}
\noindent\textbf{Federated Learning.} \ 
Federated learning (FL) enables multiple clients to collaboratively train a model in a distributed environment without sharing raw data. A representative for FL is FedAvg~\cite{fedavg}.
%
%
Unlike CL, FL typically operates under non-IID data, where client data follow different distributions, resulting in heterogeneity across clients. Such heterogeneity often leads to degraded convergence, reduced global model accuracy, and performance imbalance across clients~\cite{zhao2018federated, li2020convergence}.
%
To address these challenges, several methods~\cite{li2020fedprox,karimireddy2020scaffold,acar2021feddyn} mitigate client drift through proximity constraints, control variates, or regularized local objectives. 
However, these approaches still assume that a single global model can effectively serve all clients despite their heterogeneous data distributions~\cite{sattler2020clustered}.
Personalized federated learning (pFL) methods~\cite{apfl,pfedme,zheng2025fedcalm} alleviate this limitation by learning client-specific models through local personalization. However, they typically assume a closed-client setting, where inference data originates from training clients, limiting applicability to unseen clients or sequences.

In contrast, clustered federated learning (CFL) methods address data heterogeneity by grouping clients with similar data distributions and training specialized models for each cluster.
Early studies include CFL~\cite{sattler2020clustered} and IFCA~\cite{ghosh2020efficient}, which partition clients into clusters and jointly optimize cluster-specific models to handle heterogeneous data distributions. 
Building on this idea, several one-shot clustering approaches have been proposed, including PACFL~\cite{vahidian2023efficientpacfl}, which leverages principal angles for clustering, and FedClust~\cite{islam2024fedclust}, which constructs clusters based on model parameters. 
Beyond clustering-based approaches, HCFL~\cite{guo2025enhancing} introduces an integrated framework that combines clustering with hierarchical model aggregation, while LCFed~\cite{zhang2025lcfed} further extends this direction by jointly optimizing model partitioning and global–local aggregation.

%

\vspace{1mm}\noindent\textbf{Applications in Robot Perception.} \ 
Recently, FL has been adopted in robot perception~\cite{kou2025fedema,liu2020fedvision, poggi2024federated}.
For instance, using an exponential moving average (EMA) of global models, FedEMA~\cite{kou2025fedema} addresses temporal catastrophic forgetting in street scene semantic understanding. 
In object detection, FedVision~\cite{liu2020fedvision} enables collaborative learning by allowing multiple institutions to perform data annotation and model training locally on their images without sharing the data.
On the other hand, stereo matching leverages FL to enable models to collaboratively adapt to diverse and challenging environments in real-time, improving accuracy on resource-constrained devices without sharing raw data~\cite{poggi2024federated}. These approaches still rely on a single shared model, which makes it challenging to effectively handle the diverse domains encountered in robotic settings.
%

\vspace{1mm}\noindent\textbf{Monocular Depth Estimation.} \ 
Monocular depth estimation research has continued to develop since the introduction of CNN-based deep learning approaches. 
Following the seminal work~\cite{eigen2014silog}, which first applies CNNs to monocular depth estimation, early studies primarily focus on encoder-decoder architectures~\cite{laina2016deeper}, formulating depth estimation as a regression problem.
Subsequently, CNN-based classification approaches that discretize depth values into bins~\cite{cao2017estimating} and bin-based methods~\cite{adabins, bhat2022localbins, lee2023slabins} leveraging CNNs emerge, improving performance and robustness. 
Additionally, Transformer-based approaches~\cite{ranftl2021vision} achieve notable improvements over CNNs by leveraging the global attention mechanism of Vision Transformers.
Recently, foundation model-based approaches~\cite{yang2024depth, yang2024depthv2} demonstrate strong zero-shot generalization capabilities across diverse environments.

FedSCDepth~\cite{soares2025fedscdepth} represents one of the pioneering efforts to integrate FL with self-supervised depth estimation for autonomous vehicles.
However, evaluations primarily focus on constrained datasets (\emph{e.g.}, KITTI~\cite{geiger2013vision}, DDAD~\cite{guizilini2020packing}), which provide only limited coverage of diversity encountered in real-world scenarios. 
%
In contrast, our work systematically accounts for the non-IID conditions, including multi-platform and environmental heterogeneity, as well as bi-modal depth range spanning indoor and outdoor scenes.


\section{Method}

 In this section, we present the design of \textsc{FeDepth}. 
%
Sec.~\ref{sec:problem} formalizes the federated depth estimation task and outlines the optimization issues under robot data heterogeneity.
Sec.~\ref{sec:scenario} introduces two real-world robot heterogeneity scenarios that encompass depth range, platform, and environment shifts.
To address these challenges, Sec.~\ref{sec:fedepth} presents \textsc{FeDepth}, a new soft clustering-based FL framework that carefully models the complex relationships between clients, thereby ensuring robust training and effective cluster model updates.
%

\subsection{Problem Statement}
\label{sec:problem}

\vspace{1mm}\noindent \textbf{Problem.} \
We address a core challenge in robot perception: training a monocular depth estimation model across distributed, heterogeneous robots in an FL setup. 
In this setting, each client collects and processes data independently under varying conditions. The goal is to collaboratively learn global depth models that generalize across diverse robot platforms without centralized data collection.

\vspace{1mm}\noindent \textbf{Formal Definition.} \ 
{Let $\mathcal{X} = \mathbb{R}^{H\times W\times 3}$ denote the space of RGB images and
$\mathcal{Y} = \mathbb{R}^{H\times W}$ the space of depth maps}. Given an input image $\mathbf{x}\in\mathcal{X}$,
FL aims to learn a model $f_{\boldsymbol{w}}:\mathcal{X}\rightarrow\mathcal{Y}$
 with learnable parameter \( w \).
Training is supervised using a loss function \( \mathcal{L}(\mathbf{y},\mathbf{y}^*) \), such as pixel-wise \( \ell_1 \), \( \ell_2 \), or scale-invariant loss~\cite{eigen2014silog},
where $\mathbf{y} = f_{\boldsymbol{w}}(\mathbf{x})$ is the predicted depth map and $\mathbf{y}^*$ is the corresponding ground truth.
In this setting, data is distributed across a set of clients \( \mathcal{C} \), where each client \( c \in \mathcal{C} \) holds its own local dataset \( \mathcal{X}_c \). The global training objective is to minimize the population loss:
\begin{equation}
\begin{split}
\min_{\boldsymbol{w}} \;
F(\boldsymbol{w})&\overset{\triangle}{=}\sum_{c\in\mathcal{C}}p_{c}F_{c}(\boldsymbol{w}), \\
F_{c}(\boldsymbol{w})&\overset{\triangle}{=}
\mathbb{E}_{(\mathbf{x},\mathbf{y}^{*})\sim\mathcal{X}_{c}}
\!\Bigl[\mathcal{L}( \mathbf{y},\mathbf{y}^*)
\Bigr].
\label{eq:fl_obj}
\end{split}
\end{equation}
Here, \( F_c(\boldsymbol{w}) \) is the local objective function for client \( c \), and \( F(\boldsymbol{w}) \) is the weighted average over these with \( p_c = \frac{|\mathcal{X}_c|}{\sum_{j} |\mathcal{X}_j|} \), as formulated in FedAvg~\cite{fedavg} algorithm.

\vspace{1mm}\noindent \textbf{Challenges in Federated Depth Estimation.} \
Monocular depth estimation learns the mapping from RGB images to depth maps by modeling the underlying depth distribution. 
Yet its depth distribution depends on the acquisition setup, including the environment, robot platform, and sensor configuration. Such heterogeneity across these setups makes client data increasingly non-IID, leading to domain shifts.
To address these challenges, we pursue an FL strategy that explicitly models the domain heterogeneity across clients. Rather than assuming that a single global model can effectively capture all data distributions, we aim to uncover structure within the client population and group clients with similar characteristics. This enables the training of specialized models that are better aligned with each domain.

\begin{figure*}[t]
    \centering
    \includegraphics[width=\textwidth]{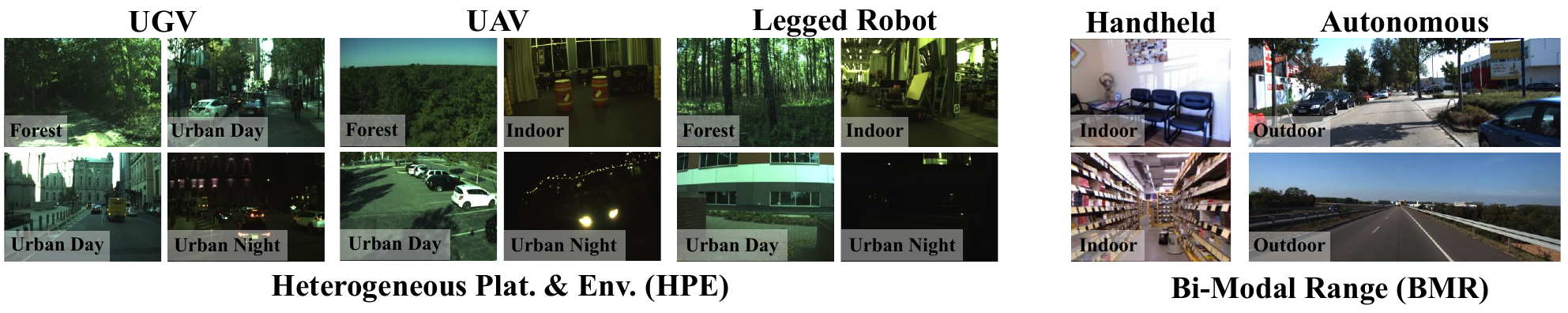}
    \caption{
    \textbf{Overview of the two scenarios for robot heterogeneity.} 
    %
    (Left) HPE introduces shifts across platforms and environments (M3ED~\cite{m3ed}).
    (Right) BMR reflects bi-modal depth distributions between indoor (NYUv2~\cite{silberman2012indoor}) and outdoor (KITTI~\cite{geiger2012we}).
    }
    \label{fig:scenarios}
\end{figure*}
\vspace{-1mm} 

\subsection{Robot Heterogeneity Scenarios}
\label{sec:scenario}

To examine the effects of heterogeneity in federated depth estimation, we introduce two real-world scenarios called HPE and BMR that reflect distinct and complementary sources of domain shift~(see Fig.~\ref{fig:scenarios}). 
These settings are designed to isolate the impact of (\emph{i})~differences in platform and environment combinations, and (\emph{ii})~differences in depth distribution across indoor and outdoor scenes.
%
Moreover, our scenarios are designed to model a spectrum of inter-client relationships, ranging from cases where clients are clearly separated into distinct distributions to cases where clients are difficult to distinguish due to overlapping distributions and multiple attributes.
%
By modeling such realistic variations, we establish a controlled yet challenging testbed for evaluating the robustness and adaptability of FL strategies under non-IID conditions. 
%
We construct the proposed two scenarios by curating existing datasets: M3ED~\cite{m3ed} for the HPE scenario, NYUv2~\cite{silberman2012indoor} and KITTI~\cite{geiger2012we} for the BMR scenario.

\begin{figure}[!t]
    \centering
    \includegraphics[width=1\linewidth]{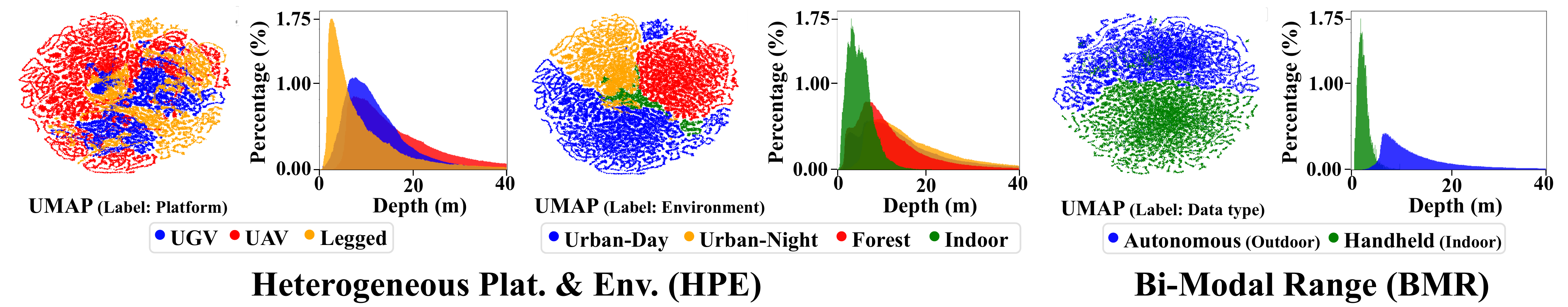}
    \caption{
    \textbf{Data characteristics of two scenarios.}
    Depth distribution and UMAP visualization~\cite{mcinnes2018umap}.}
    \label{fig:umap_and_depthRange}
\end{figure}

\vspace{1mm}\noindent \textbf{Heterogeneous Platform-Environment (HPE) Scenario.} \ 
%
{HPE Scenario contains various platforms (\emph{e.g.}, UGV, UAV, and legged robot) and environments (\emph{e.g.}, urban day/night, forest, and indoor). Each client is assigned a single continuous trajectory to reflect realistic robot deployments.
Environmental variations influence both scene geometry (\emph{e.g.}, planar indoor structures \emph{vs.} irregular outdoor foliage) and appearance characteristics (\emph{e.g.}, shape and texture distributions).
Furthermore, these platforms and environment attributes are often combinatorial and overlapping across clients (see the left of Fig.~\ref{fig:umap_and_depthRange}). For instance, Client A (Legged robot) may capture data in a `Forest' environment, while Client B (UAV) operates in the same `Forest'. While both clients share the `Forest' attribute, their data distributions diverge due to platform-specific characteristics like viewpoint and motion dynamics. This combinatorial nature and the presence of multiple, overlapping attributes make it challenging to assign a client to a single, unique group. This ambiguity undermines the clear cluster structures assumed by conventional CFL methods and leads to performance degradation under severe data heterogeneity.
}

\vspace{1mm}\noindent \textbf{Bi-Modal Range (BMR) Scenario.} \ 
The BMR scenario models a severe form of heterogeneity where clients are divided into two fundamentally distinct groups: indoor and outdoor. 
Clients consist of indoor images (\emph{e.g.}, residence interiors and market aisles) and outdoor images (\emph{e.g.}, roads and urban environments). Indoor clients observe enclosed, close-range scenes (under 10 meters), while outdoor clients capture open, far-range environments (up to 80 meters). 
This difference leads to different depth distributions, which are further amplified by discrepancies in camera parameters such as resolution, focal length, and aspect ratio. 
These combined factors result in fundamentally different supervision signals across clients (see the right of Fig.~\ref{fig:umap_and_depthRange}), allowing for the evaluation of how range heterogeneity impacts FL.
\subsection{Federated Learning for Depth Estimation}
\label{sec:fedepth}

\begin{figure*}[!t]
    \centering
    \includegraphics[width=0.99\textwidth]{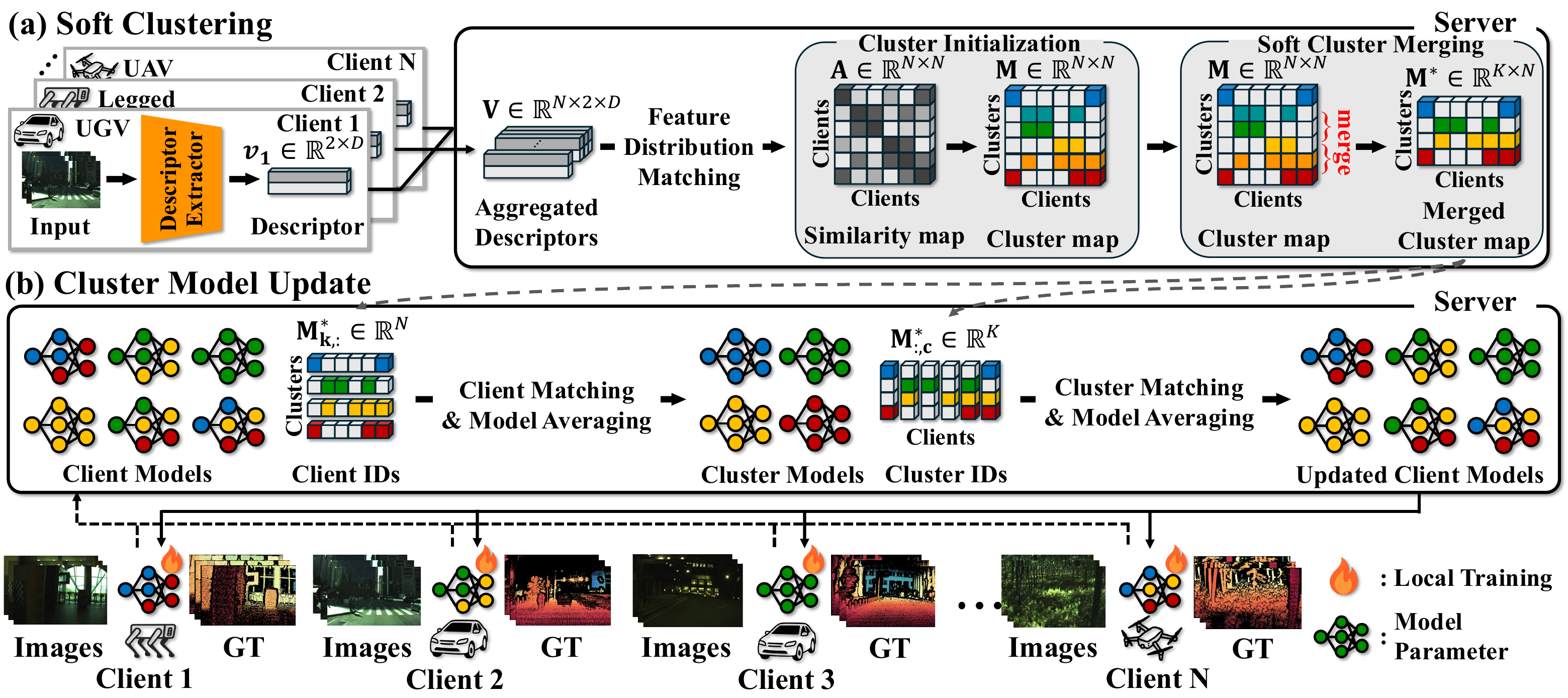}
    \vspace{-1mm} 
    \caption{{\textbf{Overall framework of \textsc{FeDepth}.} 
    (a) The Soft Clustering stage is first performed before the initial round. 
    In this stage, each client utilizes a Descriptor Extractor to generate a client descriptor summarizing its local data distribution. 
    Then, the server aggregates these descriptors and performs clustering to determine a cluster map. Second,  (b) in the Cluster Model Update stage, clients conduct iterative local training in each round. 
    At the end of each round, the server aggregates the client models, updates the cluster models, and propagates them according to the cluster map.}
    }
    \label{fig:overview}
\end{figure*}

Based on the two scenarios, $\textsc{FeDepth}$ presents a novel CFL approach to address the severe robotic heterogeneity~(see Fig.~\ref{fig:overview}).
%
$\textsc{FeDepth}$ first performs the \textit{Soft Clustering} stage, which aims to allocate each client into appropriate clusters by considering its robotic settings.
Specifically, we utilize visual descriptors, which are extracted by a pre-trained encoder to model data distribution of each client. 
Based on the pairwise similarity between these descriptors, $\textsc{FeDepth}$ constructs a soft cluster map that captures complex inter-client relationships. 
This design allows each client to participate in multiple clusters, thereby modeling partially overlapping data distributions that frequently arise in heterogeneous robotic deployments.
In the \textit{Cluster Model Update} stage, clients upload their locally-trained models, and the server performs cluster-wise model aggregation for each cluster. 
%
Finally, the server sends each client an updated model obtained by averaging the models of the clusters to which the client belongs.
In the \textit{Inference} stage, each unseen sequence is assigned to the relevant cluster models based on similarity and the clustering threshold.

\vspace{1mm}\noindent \textbf{Soft Clustering.} \ 
In robotic settings, clustering based on image distributions that capture variations in motion or environment is closely related to visual place recognition~\cite{radenovic2018fine,arandjelovic2016netvlad}. 
Inspired by this, we propose a descriptor-based soft clustering scheme that leverages the characteristics of feature distribution (see Fig.~\ref{fig:overview}a).
%
To represent the feature distribution of each client, we employ a descriptor extractor to obtain a client-level descriptor.
For each client $c$, we randomly sample a subset $\mathcal{X}'_c \subseteq \mathcal{X}_c$ of size $n_c = \min(n, |\mathcal{X}_c|)$ and extract the feature map for each image $\mathbf{x}_c^i \in \mathcal{X}'_c$:
\[
z_c^i = f_{\text{enc}}(\mathbf{x}_c^i) \in \mathbb{R}^{H'\times W'\times D},
\]
%
%
where $f_{\text{enc}}(\cdot)$ denotes the pre-trained encoder of the respective baseline models (\emph{e.g.}, AdaBins~\cite{adabins}, NeWCRFs~\cite{newcrfs}, and DCDepth~\cite{dcdepth}).
Then, to effectively compress this feature, we utilize generalized mean (GeM) pooling~\cite{radenovic2018fine}: 
\begin{equation}
g_c^i = \left(\frac{1}{H'W'} \sum_{h}\sum_{w} (z_{c}^{\,i}(h,w))^{p}\right)^{\frac{1}{p}}, \quad p>0.
\end{equation}
We aggregate $g_c^i$ into an aggregated descriptor $\mathbf{G}_c$, and compute its mean $\mu_c$ and variance $\sigma_c^2$ to
represent the feature distribution of each client. 
%
The pair $(\mu_c,\sigma_c^2)$ defines the client descriptor $v_c$, capturing both the central tendency and variability of client features, enabling robust comparison between heterogeneous clients.
The resulting descriptor is then transmitted to the server. 
On the server side, all aggregated client descriptors $\mathbf{V}=\{v_c\}_{c=1}^{N}$ are utilized to perform clustering suitable for the robotic setting. 
To assess the similarity between client domains, we approximate each descriptor with a Gaussian distribution with diagonal covariance for computational efficiency and perform feature distribution matching.
%
Although the discrepancy between two Gaussian distributions can be measured using the KL divergence, it is directional and depends on the order of the two distributions. To obtain an order-invariant measure, we employ the Jeffreys divergence~\cite{jeffreys1998theory}, a symmetric bidirectional form of the KL divergence, which is defined as follows:
\begin{equation}
A_{i,j} = D_{\text{KL}}\!\big(\mathcal{N}(\mu_i,\sigma_i^2)\,\|\,\mathcal{N}(\mu_j,\sigma_j^2)\big)
          + D_{\text{KL}}\!\big(\mathcal{N}(\mu_j,\sigma_j^2)\,\|\,\mathcal{N}(\mu_i,\sigma_i^2)\big),
\label{eq:jeffreys}
\end{equation}
where $\mathcal{N}(\cdot,\cdot)$ denotes the Gaussian distribution of each client, and $D_{KL}(\cdot||\cdot)$ is the KL divergence. 
Using Eq.~(\ref{eq:jeffreys}), we construct the similarity map $\mathbf{A} \in \mathbb{R}^{N\times N}$, where $N$ denotes the number of clients, and each entry $A_{i,j}$ represents the pairwise similarity between clients. 
However, due to robot heterogeneity, the domains of clients are intricately mixed, 
making it difficult to clearly separate clusters using this naive similarity map with simple criteria.

To handle this complexity, we employ a soft clustering strategy, which allows a client to belong to multiple clusters.
Concretely, we first generate a binary initial cluster map
$\mathbf{M} \in \{0,1\}^{N \times N}$ to group clients with high similarity based on a clustering threshold $\tau$.
This thresholding step effectively identifies strongly related clients while still allowing overlapping memberships across clusters.
Here, each row $\mathbf{M}_{k,:}$ corresponds to the set of clients assigned to cluster $k$.
By adjusting the threshold $\tau$, the clustering behavior can be dynamically controlled, 
ranging from highly overlapping soft clusters to strictly disjoint hard clusters.

Given the initial cluster map $\mathbf{M}$, 
we perform a soft cluster merging step to resolve redundant partitions, as several preliminary clusters exhibit highly correlated feature distributions.
%
For any pair of clusters $\mathbf{M}_{i,:}$ and $\mathbf{M}_{j,:}$, if one cluster is a subset of the other, the subset cluster is merged into the dominant cluster. 
The merging step iterates until convergence, resulting in the final merged cluster map $\mathbf{M}^*\in \{0,1\}^{K \times N}$, 
where $K$ denotes the number of clusters, each row $\mathbf{M}^*_{k,:}$ encodes the client IDs belonging to cluster $k$, while each column $\mathbf{M}^*_{:,c}$ encodes the cluster IDs associated with client $c$ (see Fig.~\ref{fig:cluster_viz}).



\vspace{1mm}\noindent \textbf{Cluster Model Update.} \ 
As shown in Fig.~\ref{fig:overview}b, built upon the FL formulation in Eq.~\eqref{eq:fl_obj}, \textsc{FeDepth} runs repetitive rounds of local training and aggregation. For the cluster-wise aggregation, each cluster $k$ aggregates the locally trained models from its component clients by client IDs $\mathbf{M}^*_{k,:}$ to compute the model for cluster $k$ at round $t$, \emph{i.e.}, $w_{k}^t$. 
On the client side, each client $c$ downloads the associated cluster models by cluster IDs $\mathbf{M}^*_{:,c}$ and takes the average of them to initialize the local training of the next round, \emph{i.e.}, $w_{c}^{t+1}$:
\begin{equation}
w_k^t \leftarrow \sum_{c \in \mathbf{M}^{*}_{k,:}} p_{c}^k w_c^t,
\qquad
w_c^{t+1} \leftarrow \frac{1}{|\mathbf{M}^{*}_{:,c}|}\sum_{k \in \mathbf{M}^{*}_{:,c}} w_k^t,
\label{eq:averaging}
\end{equation}
where $p_{c}^k = \frac{|\mathcal{X}_c|} {\sum_{j \in \mathbf{M}^{*}_{k,:}} |\mathcal{X}_j|}$
is the data-size weight in cluster $k$, as in FedAvg~\cite{fedavg}.

\vspace{1mm}\noindent \textbf{Inference.} \ 
At inference time, we extract the descriptor $v_{\text{test}}$ of a test sequence using the same frozen encoder as in the clustering stage. 
The similarity between the test descriptor and cluster $k$ is computed as
\begin{equation}
    S_{\text{test}}^k = \frac{1}{|\mathbf{M}^*_{k,:}|} \sum_{c \in \mathbf{M}^*_{k,:}} A_{\text{test},c},
\end{equation}
where $A_{i,j}$ denotes the Jeffreys divergence defined in Eq.~\eqref{eq:jeffreys}. 
Following the soft clustering step, we apply the same threshold $\tau$ and define the candidate cluster set as $\mathcal{K}_{\text{test}} = \{ k \mid S_{\text{test}}^k \le \tau \}$.
The prediction model is obtained by averaging the selected cluster models, consistent with Eq.~\eqref{eq:averaging}. If no cluster satisfies the threshold, the test sequence is assigned to $\arg\min_k S_{\text{test}}^k$. Cluster selection is performed once per sequence, after which the selected cluster model is used to perform depth estimation for the remaining frames of the sequence.
The full \textsc{FeDepth} procedure is summarized in the supplementary material.

\section{Experiments}
\subsection{Experimental Setup}
To validate the proposed method, we apply CL, FL, CFL, and \textsc{FeDepth} to various monocular depth estimation baselines under the HPE and BMR scenarios.
Due to the high computational cost of federated depth estimation (2--4 days per experiment using four RTX 4090 GPUs), we follow standard practice in monocular depth estimation~\cite{adabins,newcrfs,dcdepth}. Results are reported as mean over the last five rounds for stability, with standard deviations provided in the supplementary material.
In HPE, the clustering threshold $\tau$ is selected to yield an overlap ratio of 20\%. For the BMR scenario, we set $\tau$ such that the resulting cluster assignments closely approximate those of the hard clustering method to explicitly verify the effectiveness of the clustering. 

\vspace{1mm}\noindent \textbf{Datasets.} \ 
In the HPE scenario, we employ M3ED~\cite{m3ed} and construct a train/test split that maximizes coverage of platform-environment combinations. Specifically, we split 41 sequences for training and include 11 unseen sequences in the test set to ensure evaluation under diverse conditions. 
Each sequence is treated as a separate client (41 clients total), and is annotated by environment type: Indoor, Urban-Day, Urban-Night, or Forest. Full details of the split and annotation are provided in the supplementary material.  
    
In the BMR scenario, we use NYUv2~\cite{silberman2012indoor} and KITTI~\cite{geiger2012we} as two distinct data distributions. 
%
We construct 32 clients, including 15 clients from NYUv2 and 17 clients from KITTI. Each client performs training on a single dataset.
We follow the official NYUv2 train/test split and adopt the KITTI split proposed by~\cite{eigen2014silog}. 
%


\vspace{1mm}\noindent \textbf{Baselines.} \ 
 The choice of model architecture can influence the performance of FL.
Thus, we provide a comprehensive evaluation by benchmarking combinations of various depth estimation models and representative FL algorithms.
Specifically, we cover diverse depth estimation baseline architectures, such as AdaBins~\cite{adabins}, NeWCRFs~\cite{newcrfs}, DCDepth~\cite{dcdepth}, and follow the standard robotics protocol of testing generalization on unseen clients (\emph{i.e.}, data from clients not encountered during training).
%
For FL algorithms, we benchmark FedAvg~\cite{fedavg}, as well as methods designed to mitigate heterogeneity, including FedProx~\cite{li2020fedprox}, FedDyn~\cite{acar2021feddyn}, SCAFFOLD~\cite{karimireddy2020scaffold}, in addition to representative CFLs~\cite{vahidian2023efficientpacfl,islam2024fedclust}.
%
Given the focus on generalization to unseen clients in robotic settings, we include representative CFL methods designed for such scenarios. In particular, FedClust~\cite{islam2024fedclust} performs hard clustering based on model parameters, and PACFL~\cite{vahidian2023efficientpacfl} applies hard clustering using image representations.
To accommodate high-resolution robotic data, we modify PACFL by replacing raw images with our descriptor extractor, and use this adapted version as PACFL$^\dagger$.
As pFL methods~\cite{apfl,pfedme,zheng2025fedcalm} typically involve client-side adaptation at inference time, which differs from our evaluation protocol without test-time updates, we report additional comparisons in the supplementary material.

\begin{table*}[t]
\centering
  \caption{
  {\textbf{Quantitative evaluation on the HPE scenario.} Methods are categorized into CL, FL methods (FedAvg~\cite{fedavg}, FedProx~\cite{li2020fedprox}, FedDyn~\cite{acar2021feddyn}, SCAFFOLD~\cite{karimireddy2020scaffold}), CFL approaches (FedClust~\cite{islam2024fedclust}, PACFL~\cite{vahidian2023efficientpacfl}), and the proposed framework (Ours). PACFL$^\dagger$ denotes the modified version of PACFL that uses our descriptor extractor. The best result per metric for each baseline is \textbf{bold}, and the second best is \underline{underlined.}}
  } 
\resizebox{0.99\linewidth}{!}{%
  \centering
  { \small
    \begin{tabular}{c|>{\centering\arraybackslash}m{2.1cm}|cccc|ccc}
    \toprule
    \textbf{Baseline}   & \textbf{Method} & $Abs\ Rel\downarrow$ & $Sq\ Rel\downarrow$ & $RMSE\downarrow$ & $RMSE_{\log}\downarrow$ & $\delta < 1.25\uparrow$ & $\delta < 1.25^2\uparrow$ & $\delta < 1.25^3\uparrow$ \\
    \toprule

    
    \multirow[c]{9}{*}{AdaBins~\cite{adabins}}
                            &    CL          &          0.222         &          1.477         &         3.552          &         0.254          &         0.733          &         0.887          &         0.941          \\
    \cmidrule{2-9}
                            &    FedAvg      &   \underline{0.434}    &  \underline{2.990}     &         6.959          &         0.542          &         0.369          &         0.580          &         0.729          \\
                            &    FedProx     &          0.439         &          3.029         &    \underline{6.933}   &         0.533          &         0.369          &         0.585          &         0.737          \\
                            &    FedDyn      &          0.440         &          3.063         &         7.027          &         0.551          &         0.367          &         0.573          &         0.721          \\
                            &    SCAFFOLD    &          0.631         &          6.009         &         9.664          &         0.749          &         0.225          &         0.432          &         0.603          \\
    \cmidrule{2-9}                            
                            &    FedClust    &          0.444         &          3.263         &         7.521          &         0.566          &         0.277          &         0.550          &         0.733          \\
                            &$\text{PACFL}^\dagger$&    0.602         &          6.447         &         7.348          &   \underline{0.509}    &    \underline{0.372}   &   \underline{0.605}    &   \underline{0.768}    \\
                            &    Ours        &     \textbf{0.318}     &      \textbf{2.127}    &    \textbf{ 5.354 }    &     \textbf{0.364}     &      \textbf{0.515}    &      \textbf{0.770}    &     \textbf{0.897}     \\

    \midrule\midrule
    

    \multirow[c]{9}{*}{NeWCRFs~\cite{newcrfs}}
                            &    CL          &          0.166         &          0.994         &         2.987          &         0.190          &         0.827          &         0.935          &         0.968          \\
    \cmidrule{2-9}
                            &    FedAvg      &          0.366         &          1.994         &         4.475          &         0.406          &         0.504          &         0.717          &         0.834          \\
                            &    FedProx     &          0.364         &          1.983         &         4.479          &         0.405          &         0.504          &         0.718          &         0.834          \\
                            &    FedDyn      &          0.363         &          1.957         &   \underline{4.473}    &         0.404          &         0.505          &         0.719          &         0.834          \\
                            &    SCAFFOLD    &          0.356         &          2.060         &         4.569          &         0.376          &         0.498          &         0.726          &         0.864          \\
    \cmidrule{2-9}
                            &    FedClust    &          0.487         &         4.325          &         7.760          &         0.615          &         0.241          &         0.507          &         0.701          \\
                            &$\text{PACFL}^\dagger$&\underline{0.318} &  \underline{1.934}     &         4.265          &  \underline{0.327}     &  \underline{0.540}     &  \underline{0.791}     &  \underline{0.928}     \\
                            &    Ours       &     \textbf{0.249}      &    \textbf{1.750}      &    \textbf{3.838}      &     \textbf{0.261}     &    \textbf{0.703}      &     \textbf{0.892}     &    \textbf{0.954}      \\

    \midrule\midrule

    \multirow[c]{9}{*}{DCDepth~\cite{dcdepth}}
                            &    CL           &          0.159         &          0.962         &         2.916          &         0.183          &         0.835          &         0.939          &         0.971          \\
    \cmidrule{2-9}
                            &    FedAvg       &          0.351         &         2.077          &         4.640          &         0.348          &         0.493          &         0.766          &         0.902          \\
                            &    FedProx      &          0.354         &         2.152          &         4.641          &         0.348          &         0.492          &         0.769          &         0.902          \\
                            &    FedDyn       &          0.352         &         2.109          &         4.699          &         0.349          &         0.491          &         0.767          &         0.903          \\
                            &    SCAFFOLD     &          0.329         &         2.050          &         4.880          &         0.350          &         0.472          &         0.759          &         0.914          \\
    \cmidrule{2-9}
                            &    FedClust     &          0.349         &         3.132          &         6.423          &         0.434          &         0.498          &         0.712          &         0.815          \\
                            &$\text{PACFL}^\dagger$& \underline{0.321} &  \underline{1.976}     &     \textbf{4.338}     &   \underline{0.325}    &   \underline{0.537}    &   \underline{0.796}    &   \underline{0.924}    \\
                            &    Ours         &     \textbf{0.293}     &     \textbf{1.872}     &   \underline{4.469}    &     \textbf{0.310}     &     \textbf{0.547}     &     \textbf{0.840}     &     \textbf{0.941}     \\

    \bottomrule
  \end{tabular}
}
  }
  \normalsize

  \label{tab:depth-comparison-m3ed}
\end{table*}

\subsection{Evaluations}
\begin{figure*}[t]
    \centering
    \includegraphics[width=0.99\linewidth]{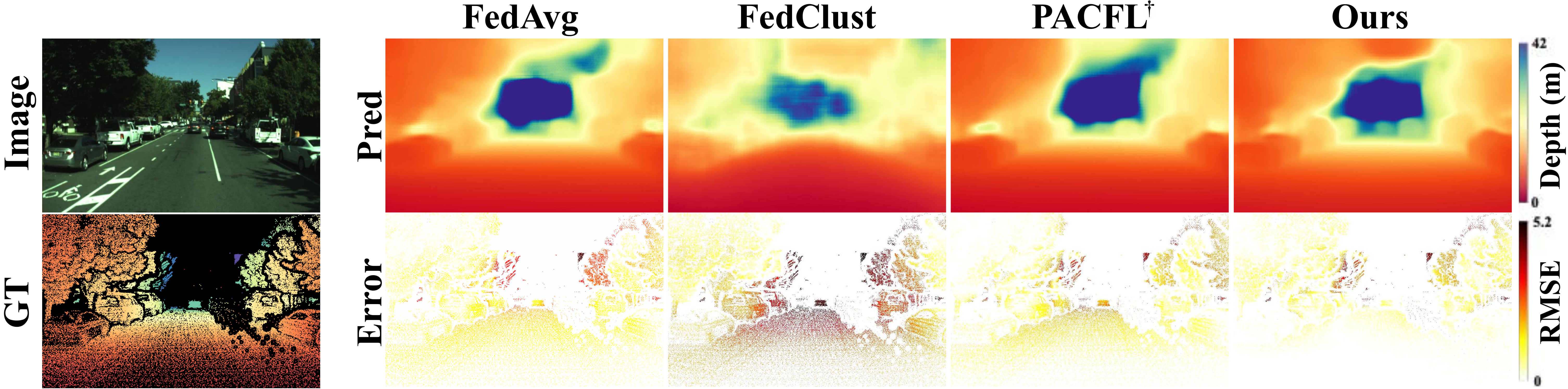} 
    \vspace{-1mm}
    \caption{
        \textbf{Qualitative results for DCDepth on the HPE scenario.} (Left) Input image and Ground Truth (GT). 
        (Right) Prediction maps (1st row) and error (RMSE) maps (2nd row) for our method (Ours) and the baselines (FedAvg, FedClust, PACFL$^\dagger$).
        %
    }
    \label{fig:Qualit_result}
\end{figure*}

\begin{figure}[t!]
    \centering
    \includegraphics[width=0.99\linewidth]{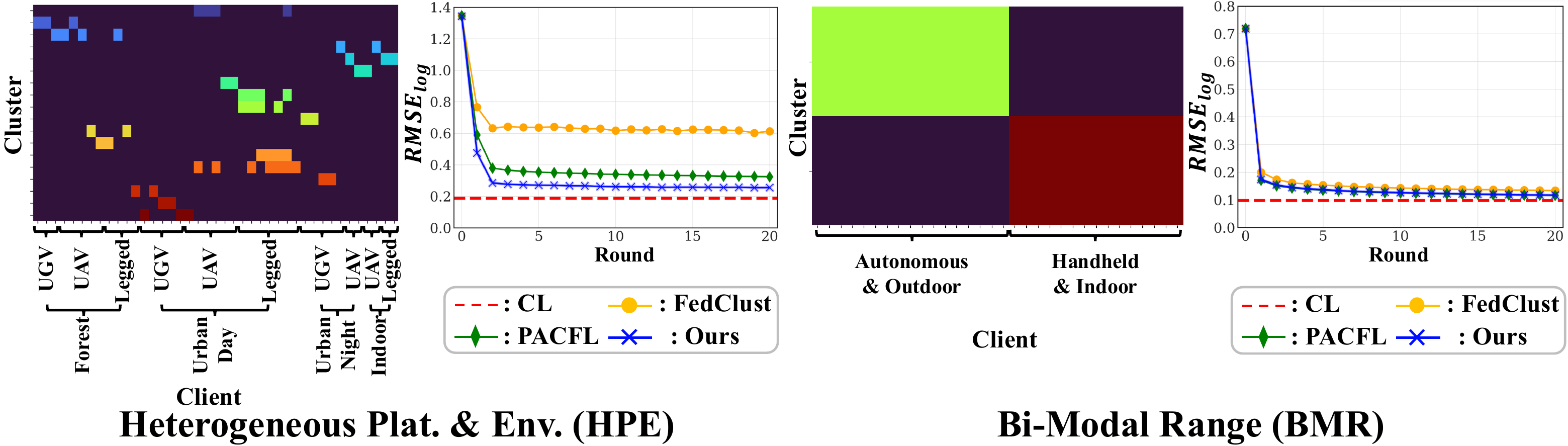}
    \vspace{-1mm}
    \caption{\textbf{Cluster map and convergence graph}. 
        %
    Results for NeWCRFs~\cite{newcrfs} in the HPE and BMR scenarios. Colors indicate cluster assignments, with each column representing a client. The convergence graphs report $RMSE_{log}$ over communication rounds.
    }
    \vspace{-3mm}
    \label{fig:cluster_viz}
\end{figure}

\noindent \textbf{HPE Scenario.} 
Due to the robot-driven heterogeneity in the HPE scenario, FL algorithms exhibit performance degradation compared to CL, even those explicitly designed to handle non-IID data (\emph{e.g.}, FedProx, FedDyn, and SCAFFOLD), as shown in Tab.~\ref{tab:depth-comparison-m3ed}.
While feature-based CFL methods like PACFL$^\dagger$ form reasonable clusters and thereby show improvements over standard FL, their performance is inherently limited by the hard clustering representation when dealing with complex, overlapping distributions. Moreover, this limitation becomes more pronounced under inaccurate cluster assignments, as observed with AdaBins trained via FedClust.
In contrast, \textsc{FeDepth} effectively represents the complex and overlapping data distributions inherent in robotic perception by utilizing soft cluster assignments. 
%
Consequently, as shown in Tab.~\ref{tab:depth-comparison-m3ed}, and Fig.~\ref{fig:Qualit_result}, \textsc{FeDepth} consistently outperforms all other FL baselines across all evaluated depth estimation architectures. 
The convergence curves in Fig.~\ref{fig:cluster_viz} show faster and more stable convergence.
This demonstrates that \textsc{FeDepth} provides an effective and model-agnostic framework for FL under complex robot-driven heterogeneity.



\begin{table*}[t]
\centering
  \caption{
  {\textbf{Quantitative evaluation on the BMR scenario.}} Additional quantitative results are available in the supplementary material.
  }
\resizebox{0.99\linewidth}{!}{%
  \centering
  { \small
    \begin{tabular}{c|>{\centering\arraybackslash}m{1.6cm}|cccc|ccc}
    \toprule
    \textbf{Baseline} & \textbf{Method} & $Abs\ Rel\downarrow$ & $Sq\ Rel\downarrow$ & $RMSE\downarrow$ & $RMSE_{\log}\downarrow$ & $\delta < 1.25\uparrow$ & $\delta < 1.25^2\uparrow$ & $\delta < 1.25^3\uparrow$ \\
    \toprule

    \multirow{4}{*}{AdaBins~\cite{adabins}}
                            &    CL            &        0.166        &        0.369        &       2.073         &       0.191         &        0.768        &        0.945         &        0.984         \\
    \cmidrule{2-9}
                            &  FedAvg          &        0.357        &        1.153        &       3.329         &       0.409         &        0.211        &        0.673         &        0.913          \\ 
                            &  PACFL$^\dagger$ &    \textbf{0.124}   &    \textbf{0.263}   &    \textbf{2.022 }   &    \textbf{0.164}   &   \textbf{0.831}    &   \textbf{0.970}    &   \textbf{0.994}     \\ 
                            &  Ours            &  \underline{0.126}  &  \underline{0.283}  &   \underline{2.150}  &  \underline{0.167}  &  \underline{0.827}  &  \underline{0.968}  &   \textbf{0.994}      \\ 

    \midrule\midrule
    

    \multirow{4}{*}{NeWCRFs~\cite{newcrfs}}
                            &    CL            &        0.077        &        0.100        &        1.209         &        0.100        &        0.950       &        0.994         &        0.998        \\
    \cmidrule{2-9}
                            &  FedAvg          &        0.323        &        1.048        &        3.094         &        0.344        &        0.191       &        0.833         &        0.983     \\ 
                            &  PACFL$^\dagger$ &   \textbf{0.087}    &    \textbf{0.145}   &   \underline{1.501}  &   \textbf{0.117}    &   \textbf{0.921}   &    \textbf{0.991}    &     \textbf{0.999}     \\ 
                            &  Ours            &  \underline{0.089}  &    \textbf{0.145}   &    \textbf{ 1.499 }   &  \underline{0.118}  &   \textbf{0.921}   &    \textbf{0.991}    &     \textbf{0.999}   \\ 

    \midrule\midrule






    \multirow{4}{*}{DCDepth~\cite{dcdepth}}
                            &    CL            &        0.074        &        0.101        &        1.246         &        0.099        &        0.952       &        0.994         &        0.998        \\
    \cmidrule{2-9}
                            &  FedAvg          &        0.158        &        0.265        &        1.789         &        0.183        &        0.801       &        0.969         &        0.996 \\ 
                            &  PACFL$^\dagger$ &   \textbf{0.082}   &   \underline{0.130}  &    \textbf{1.428}    &  \underline{0.109}  &   \textbf{0.933}   &    \textbf{0.993}    &    \textbf{0.999}   \\ 
                            &  Ours            &   \textbf{0.082}   &     \textbf{0.127}   &  \underline{1.430}   &    \textbf{0.108}   &   \textbf{0.933}   &    \textbf{0.993}    &    \textbf{0.999}  \\ 
    
    \bottomrule
  \end{tabular}
}
  } \normalsize
  \label{tab:depth-comparison-nyu-kitti}
\vspace{-1mm}
\end{table*}

\vspace{1mm}\noindent \textbf{BMR Scenario.} \ 
The BMR scenario consists of two clearly separated and non-overlapping domains (indoor/outdoor), providing an ideal setting for conventional hard clustering methods such as PACFL$^\dagger$. 
As shown in Tab.~\ref{tab:depth-comparison-nyu-kitti}, FedAvg still suffers severe performance degradation, confirming that even simple domain heterogeneity is a challenge. 
In particular, as reported in prior work~\cite{bhat2023zoedepth}, bin-based methods such as AdaBins may struggle even under CL. Training a depth distribution (bins) with a single model on a mixed dataset with clearly distinct depth ranges (\emph{e.g.}, 10m vs 80m) disrupts the learned bin distribution.

In contrast, CFL mitigates this issue by grouping similar clients. Our results show that \textsc{FeDepth} is not limited to overlapping domains and can naturally form well-defined clusters when data distributions are clearly separated.
For NeWCRFs and DCDepth, near-binary cluster maps are formed (see Fig.~\ref{fig:cluster_viz}), achieving performance comparable to PACFL$^\dagger$.
For AdaBins, feature representations are less sharply separable, preventing a perfectly binary split. Consequently, although FeDepth still outperforms CL, a minor performance gap remains compared to PACFL$^\dagger$. Additional results and cluster visualizations are provided in the supplementary material.

\begin{table*}[t!]
\centering
\caption{\textbf{Ablation study.}
All results are reported using the $Abs\ Rel\downarrow$ metric based on NeWCRFs~\cite{newcrfs} under the HPE scenario. 
(a) compares hard and soft clustering under the same threshold $\tau$, 
(b) evaluates sensitivity to overlap ratios, and 
(c) reports results across train–test domain splits based on ground-truth labels.
}
\vspace{-4mm}
\small
\begin{subtable}[t]{0.29\textwidth}
\centering
\caption{Hard vs. Soft clustering}
\vspace{-2mm}
\begin{adjustbox}{width=\linewidth}
\begin{tabular}{c|cc}
\toprule
\textbf{Cluster type} & \textbf{Hard} & \textbf{Soft} \\
\midrule
$Abs\ Rel\downarrow$ & 0.264 & \textbf{0.249} \\
\bottomrule
\end{tabular}
\end{adjustbox}
\label{tab:ablation_soft_valid}
\end{subtable}
\hfill
\begin{subtable}[t]{0.673\textwidth}
\centering
\caption{Overlap sensitivity}
\vspace{-2mm}
\begin{adjustbox}{width=\linewidth}
\begin{tabular}{c|ccccccc|c}
\toprule
\textbf{Overlap (\%)} 
& \textbf{0} & \textbf{10} & \textbf{20} & \textbf{30} 
& \textbf{40} & \textbf{50} & \textbf{100} & \textbf{FedAvg} \\
\midrule
$Abs\ Rel\downarrow$ 
& 0.335 & 0.253 & \textbf{0.249} & 0.257 
& 0.295 & 0.300 & 0.365 & 0.366 \\
\bottomrule
\end{tabular}
\end{adjustbox}
\label{tab:ablation_overlap}
\end{subtable}

\begin{subtable}[t]{0.99\textwidth}
\centering
\caption{Domain heterogeneity}
\vspace{-2mm}
\begin{adjustbox}{width=\linewidth}
\begin{tabular}{c|c|c|ccc|cccc|c}
    \toprule
    \multirow[c]{2}{*}{\textbf{Type}} & \multirow[c]{2}{*}{\textbf{Method}} & \multirow[c]{2}{*}{\textbf{Train set}} & \multicolumn{8}{c}{\textbf{Test set}} \\
    \cline{4-6} \cline{7-10} \cline{10-11} & & & UGV & UAV & Legged & Forest & Urban-day & Urban-night & Indoor & All\\
    \toprule
    \multirow{4}{*}{All}  &     CL        &  All        &       0.122       &        0.158       &        0.261      &        0.091       &        0.112          &        0.280          &       0.284         &       0.167         \\  	 	 	 	 	 	 
    \cmidrule{2-11}       &  FedAvg       &  All        &       0.166       &        0.372       &        0.700      &        0.298       &        0.358          &        0.356          &       0.659         &       0.366         \\   	 	 	 	 	 	 
                          &PACFL$^\dagger$&  All        &       0.174       & \underline{0.305}  &        0.580      &  \underline{0.236} &        0.338          &        0.414          &       0.309         &  \underline{0.318}  \\
                          &     Ours      &  All        &\underline{0.148}  &  \textbf{0.262}    & \underline{0.411} &    \textbf{0.203}  &  \textbf{0.185}       &        0.420          &  \textbf{0.275}     &   \textbf{0.249}    \\
    \midrule
    
    \multirow{3}{*}{Plat.}&    FedAvg     & UGV        &   \textbf{0.139}  &        0.514       &        0.996      &        0.493       &        0.366          &   \underline{0.318}   &       0.663         &       0.471         \\  
                          &    FedAvg     & UAV        &       0.295       &        0.640       &        1.358      &        0.790       &        0.549          &        0.471          &       0.610         &       0.662         \\ 	 
                          &    FedAvg     &Legged      &       0.458       &        0.645       &   \textbf{0.265}  &        0.407       &        0.413          &        0.483          &       0.782         &       0.484         \\ 
    \midrule
    
    \multirow{4}{*}{Env.} &   FedAvg      & Forest     &       0.269       &        0.383       &        0.530      &        0.246       &        0.351          &        0.381          &       0.764         &       0.370         \\ 
                          &   FedAvg      & Urban-day  &       0.191       &        0.432       &        0.696      &        0.415       & \underline{0.292}     &  \textbf{0.301}       &       0.680         &       0.394         \\  	
                          &   FedAvg      & Urban-night&       0.204       &        0.502       &        0.943      &        0.527       &        0.393          &        0.330          &       0.657         &       0.481         \\ 
                          &   FedAvg      & Indoor     &       0.709       &        0.823       &        2.012      &        1.297       &        1.237          &        0.764          &  \underline{0.279}  &       1.069         \\   
    \bottomrule
  \end{tabular}
  \vspace{-3mm}
\end{adjustbox}
\label{tab:ablation_domain}
\end{subtable}
\vspace{-2mm}
\label{tab:ablation_study}
\end{table*}

\vspace{1mm}\noindent \textbf{Ablation Study.} \
We conduct ablation experiments to evaluate the proposed clustering strategy (Tab.~\ref{tab:ablation_study}).
In Tab.~\ref{tab:ablation_soft_valid}, we replace our soft clustering with the hard clustering~\cite{day1984efficient} used in PACFL$^\dagger$, resulting in a hard clustering setting under the same threshold $\tau$. 
Under the same setup, soft clustering performs better than hard clustering.
This suggests that data distributions in robotic environments are continuous and overlapping, making flexible clustering more suitable than hard partitioning.
%
%

Tab.~\ref{tab:ablation_overlap} reports a sensitivity study on the overlap ratio in soft clustering, defined as the fraction of clients assigned to multiple clusters. As shown in the table, a small overlap limits knowledge sharing among related domains, whereas a large overlap increases intra-cluster heterogeneity. In the extreme case of 100\% overlap, all clients collapse into a single cluster, making the setting equivalent to FedAvg. As a result, we adopt an overlap ratio of 20\%, which yields the best performance in our experiments.
%
%

Tab.~\ref{tab:ablation_domain} compares our approach with clustering based on ground-truth labels, where clients are grouped by platform or environment and trained separately before evaluation on the same test set. 
Although grouping with known labels represents a well defined cluster, unstable performance is observed.
This indicates that grouping clients solely by known labels does not necessarily yield optimal performance. 
In contrast, \textsc{FeDepth} performs distribution-aware clustering that captures finer-grained similarities between clients, leading to consistently better results. 
Additional results and cluster visualizations are provided in the supplementary material.

%
%

\section{Conclusion}
We have proposed \textsc{FeDepth}, a clustered federated learning (CFL) framework that addresses the non-IID challenges of depth estimation for robots in the real world. 
By utilizing soft cluster assignments, \textsc{FeDepth} alleviates the performance degradation of FL that is commonly observed under robot-driven heterogeneity. To enable systematic analysis, we design and construct two realistic and previously unexplored robot heterogeneity scenarios that highlight the impact of platform, environment, and depth range variations. These scenarios, together with our framework, serve as a practical benchmark and foundation for applying FL in diverse real-world robotic systems.
We believe that \textsc{FeDepth} represents an important step toward scalable perception across heterogeneous robots, bringing FL closer to practical deployment. 
%

\section*{Acknowledgements}
This work was supported by the Institute of Information \& Communications Technology Planning \& Evaluation (IITP) grant funded by the Korea government (MSIT) (No.RS-2020-II201336, Artificial Intelligence Graduate School Program (UNIST);
No.RS-2025-25442824, AI Star Fellowship Program (UNIST);
No.RS-2022-II220907, Development of AI Bots Collaboration Platform and Self-organizing), 
and by the National Research Foundation of Korea~(NRF) grant funded by the Korea government~(MSIT) (No.RS-2024-00457065).

%
%
\bibliographystyle{splncs04}
\bibliography{main}

\newpage
\appendix



\newcommand{\makesupptitle}[1]{%
  \clearpage
  \begin{center}%
    {\Large \bfseries\boldmath #1 \par}\vskip .8cm
    {\large Supplementary Material \par}\vskip .4cm
  \end{center}%
  \vskip .6cm
  
  \setcounter{section}{0}
  \setcounter{equation}{0}
  \setcounter{figure}{0}
  \setcounter{table}{0}
  \renewcommand{\thesection}{A\arabic{section}} 
  \renewcommand{\theequation}{S\arabic{equation}} 
  \renewcommand{\thefigure}{S\arabic{figure}} 
  \renewcommand{\thetable}{S\arabic{table}} 
}


\clearpage
\setcounter{page}{1}
\makesupptitle{FeDepth: Federated Learning for Depth Estimation under Robot Heterogeneity}

\renewcommand{\thefigure}{A\arabic{figure}}
\renewcommand{\thetable}{A\arabic{table}}
\renewcommand{\theequation}{A\arabic{equation}}
\renewcommand{\thesection}{A\arabic{section}}

\section*{Overview}
This supplementary material provides additional details and experimental results to complement the main paper.
\begin{itemize}
    \item \textbf{Sec.~\ref{sec:method_detail}} presents the full \textsc{FeDepth} pipeline together with the pseudo-code for descriptor extraction, soft clustering, and cluster assignment at inference.
    \item \textbf{Sec.~\ref{sec:additional_experiments}} reports extended quantitative results, including standard deviations for the HPE and BMR scenarios, additional ablation results, and comparisons with personalized federated learning methods.
    \item \textbf{Sec.~\ref{sec:clustering_result}} provides a detailed analysis of the clustering behavior under heterogeneous scenarios.
    \item \textbf{Sec.~\ref{sec:visualization_result}} presents additional qualitative results and convergence graphs for different baselines.
    \item \textbf{Sec.~\ref{sec:implementation_detail}} describes the preprocessing steps, dataset construction, and implementation details to facilitate reproducibility.
\end{itemize}

\section{Method Details}
\label{sec:method_detail}
\begin{figure}
    \centering
    \includegraphics[width=0.99\linewidth]{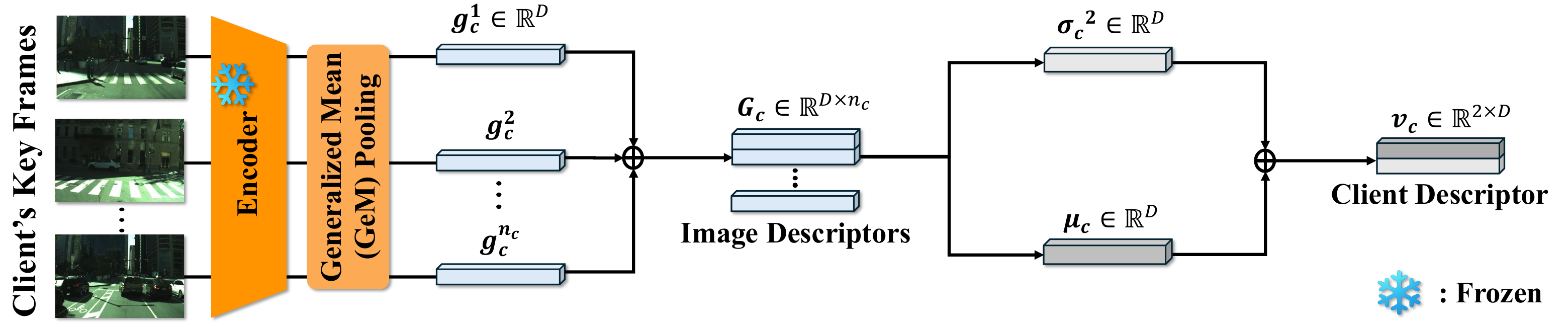}    
    \caption{\textbf{Descriptor Extractor.} Keyframes are encoded by a frozen backbone and aggregated via GeM pooling. Their mean and variance define the client descriptor.}
    \label{fig:descriptor}
    \vspace{-8mm}
\end{figure}

\vspace{1mm} \noindent \textbf{Pseudo-code of \textsc{FeDepth}.} \ 
To clarify the overall workflow of the proposed method, we provide the pseudo-code of \textsc{FeDepth} in this section.
Before federated training, each client extracts a descriptor using Alg.~\ref{alg:descriptor} (see Fig.~\ref{fig:descriptor}).
%
The server then collects the descriptors extracted by the clients and performs soft clustering using Alg.~\ref{alg:clustering}. In particular, the algorithm constructs a similarity map using the Jeffreys divergence~\cite{jeffreys1998theory}, initializes a binary cluster map, 
merges redundant clusters based on subset relations, and reassigns singleton clusters to their nearest clusters for stability.
%
The cluster-wise federated training procedure then follows Alg.~\ref{alg:fedepth}, where clients independently update their local models and the server aggregates updates within each cluster.
%
At inference time, each test sequence is assigned to the most relevant cluster models according to Alg.~\ref{alg:cluster_inference} based on descriptor similarity, allowing multiple clusters to be selected in a soft manner rather than enforcing a single hard assignment, thereby improving generalization to unseen domains.


\newpage
\vspace{-4mm}
\begin{center}
\resizebox{0.9\linewidth}{!}{%
\begin{minipage}{\linewidth}
\begin{algorithm}[H]
\caption{Descriptor Extraction}
\label{alg:descriptor}
\begin{algorithmic}[1]
\Require Local dataset $\mathcal{X}_c$, number of sampled keyframes $n$

\State $n_c \leftarrow \min(n, |\mathcal{X}_c|)$
\State $\mathcal{X}'_c \leftarrow$ sample $n_c$ keyframes from $\mathcal{X}_c$

\For{each image $x_c^i \in \mathcal{X}'_c$}
    \State $z_c^i \leftarrow f_{enc}(x_c^i)$ 
    \State $g_c^i \leftarrow 
    \left(
    \frac{1}{H'W'} 
    \sum_h \sum_w (z_c^i(h,w))^p
    \right)^{\frac{1}{p}}$ 
\EndFor

\State $\mu_c \leftarrow \frac{1}{n_c}\sum_{i=1}^{n_c} g_c^i$
\State $\sigma_c^2 \leftarrow \frac{1}{n_c}\sum_{i=1}^{n_c} (g_c^i-\mu_c)^2$

\State \Return $v_c=(\mu_c,\sigma_c^2)$
\end{algorithmic}
\end{algorithm}
\end{minipage}
}
\end{center}

\vspace{-7mm}
\begin{center}
\resizebox{0.9\linewidth}{!}{%
\begin{minipage}{\linewidth}
\begin{algorithm}[H]
\caption{Soft Clustering}
\label{alg:clustering}
\begin{algorithmic}[1]
\Require Descriptor set $\mathbf{V}$, threshold $\tau$
\For{each pair of clients $(i,j)$}
    \State Construct Gaussian distribution from descriptor $v_i$:
    \State $\mathcal{N}_i = \mathcal{N}(\mu_i, \sigma_i^2)$
    \State Compute similarity map $\mathbf{A} \in \mathbb{R}^{N\times N}$ using Jeffreys divergence:
    \State $A_{i,j} \gets D_{\text{KL}}(\mathcal{N}_i \parallel \mathcal{N}_j) + D_{\text{KL}}(\mathcal{N}_j \parallel \mathcal{N}_i)$
\EndFor
\State Initialize binary cluster map $\mathbf{M} $:
\State $\mathbf{M} \gets \mathbb{I}( \mathbf{A} \le \tau ) \in \{0,1\}^{N \times N}$
\Statex
\Repeat
    \For{each pair of clusters $k,\ l$}
        \If{$\mathbf{M}_{k,:} \subseteq \mathbf{M}_{l,:}$ \textbf{ or } $\mathbf{M}_{l,:} \subseteq \mathbf{M}_{k,:}$}
            \State Merge $(\mathbf{M}_{k,:}, \mathbf{M}_{l,:})$
        \EndIf
    \EndFor
\Until the merge is finished
\Statex
\For{each singleton cluster $k_{\text{single}}$ where $|\mathbf{M}_{k_{\text{single}},:}|=1$ }
    \State Compute the mean distance to each cluster:
    \For{each cluster k where $k \neq k_{\text{single}}$}
        \State $\bar{A}(k) \gets \frac{1}{|\mathbf{M}_{k,:}|} \sum_{c \in \mathbf{M}_{k,:}} A_{k_{\text{single}},c}$
    \EndFor
    \State $t \gets \arg\min_k \bar{A}(k)$
    \State Merge $(\mathbf{M}_{k_{\text{single}},:}, \mathbf{M}_{t,:})$
\EndFor
\State \Return final merged cluster map $\mathbf{M}^{*}$
\end{algorithmic}
\end{algorithm}
\end{minipage}
}
\end{center}

\begin{center}
\resizebox{0.9\linewidth}{!}{%
\begin{minipage}{\linewidth}
\begin{algorithm}[H]
\caption{\textsc{FeDepth} Overview}
\label{alg:fedepth}
\centering
\begin{algorithmic}[1]
\Require Initial model $w^0$, client set $\mathcal{C}$, communication rounds $T$, threshold $\tau$

\State \textcolor{blue}{\textit{\textbf{\# Stage 1}: Client Descriptor Extraction}}
\For{each client $c \in \mathcal{C}$ \textbf{in parallel}}
    \State $v_c \gets \textit{Descriptor Extraction}(c)$ \textcolor{blue}{\Comment{Alg.~\ref{alg:descriptor}}}
\EndFor
\State $\mathbf{V} \gets \{v_c\}_{c\in\mathcal{C}}$

\Statex
\State \textcolor{blue}{\textit{\textbf{\# Stage 2}: Descriptor-based Soft Clustering}}
\State $\mathbf{M}^* \gets \textit{Soft Clustering}(\mathbf{V}, \tau)$ \textcolor{blue}{\Comment{Alg.~\ref{alg:clustering}}}

\Statex
\State \textcolor{blue}{\textit{\textbf{\# Stage 3}: Cluster Model Update}}
\For{$t = 1$ to $T$}

    \For{each client $c \in \mathcal{C}$ \textbf{in parallel}}

        \State initialize local model $w_c^t$:
        \State $w_c^t \gets
        \begin{cases}
        w^0 & t=1 \\
        \frac{1}{|\mathbf{M}^*_{:,c}|}\sum_{k\in\mathbf{M}^*_{:,c}} w_k^{t-1} & \text{otherwise}
        \end{cases}$

        \State $w_c^{t+1} \gets \textsc{ClientUpdate}(c, w_c^t)$

    \EndFor

    \For{each cluster $k$}
        \State $w_k^{t+1} \gets \sum_{c\in\mathbf{M}^*_{k,:}} p_c^k w_c^{t+1}$, \quad $p_c^k=\frac{|\mathcal{X}_c|}{\sum_{j\in\mathbf{M}^*_{k,:}}|\mathcal{X}_j|}$
    \EndFor

\EndFor
\Statex
\Procedure{ClientUpdate}{$c,\ w_c$}
        \State Split $\mathcal{X}_c$ into batches $\mathcal{B}$
        \For{each local epoch $i = 1$ to $E$}
            \For{batch $b \in \mathcal{B}$}
                \State $w_c \gets w_c - \eta \nabla\ell(w_c; b)$
            \EndFor
        \EndFor
        \State \Return $w_c$
    \EndProcedure
\end{algorithmic}
\end{algorithm}
\end{minipage}
}
\end{center}
\vspace{-8mm}

\begin{center}
\resizebox{0.9\linewidth}{!}{%
\begin{minipage}{\linewidth}
\begin{algorithm}[H]
\caption{Cluster Assignment at Inference}
\label{alg:cluster_inference}
\begin{algorithmic}[1]
\Require Test image set $\mathcal{X}_{\text{test}}$, cluster map $\mathbf{M}^*$, client descriptors $\mathbf{V}$, threshold $\tau$

\State Extract descriptor for the test sequence:
\State $v_{\text{test}} \gets \textit{Descriptor Extractor}(\mathcal{X}_{\text{test}})$ \textcolor{blue}{\Comment{Alg.~\ref{alg:descriptor}}}

\For{each cluster $k$}
    \For{each client $c$ such that $M^*_{k,c} = 1$}
        \State $A_{\text{test},c} \gets D_{KL}(\mathcal{N}_{\text{test}}\|\mathcal{N}_c) + D_{KL}(\mathcal{N}_c\|\mathcal{N}_{\text{test}})$
    \EndFor
    \State $S_{\text{test}}^k \gets \frac{1}{|M^*_{k,:}|}\sum_{c\in M^*_{k,:}} A_{\text{test},c}$
\EndFor

\State Candidate cluster set:
\State $\mathcal{K} \gets \{k \mid S_{\text{test}}^k \le \tau\}$

\State Compute prediction model:
\State $w_{\text{test}} \gets
\begin{cases}
\sum_{k\in\mathcal{K}} p_k w_k,
\quad
p_k=\dfrac{{S_{\text{test}}^{k}}^{-1}}
{\sum_{j\in\mathcal{K}} {S_{\text{test}}^{j}}^{-1}}
& \textbf{if } \mathcal{K}\neq\emptyset \\
w_{k^*},\; k^*=\arg\min_k S_{\text{test}}^{k}
& \textbf{otherwise}
\end{cases}$

\State \Return $w_{\text{test}}$
\end{algorithmic}
\end{algorithm}
\end{minipage}
}
\end{center}

\begin{table}
\centering
\rotatebox{90}{
\begin{minipage}{0.98\textheight}
\centering
  \caption{
  {\textbf{Quantitative evaluation on the HPE scenario.} }
  } 
\resizebox{\linewidth}{!}{%
\setlength{\tabcolsep}{5pt}
  \centering
  { \small
    \begin{tabular}{c|>{\centering\arraybackslash}m{2cm}|cccc|ccc}
    \toprule
    \textbf{Baseline} & \textbf{Method} & $Abs\ Rel\downarrow$ & $Sq\ Rel\downarrow$ & $RMSE\downarrow$ & $RMSE_{\log}\downarrow$ & $\delta < 1.25\uparrow$ & $\delta < 1.25^2\uparrow$ & $\delta < 1.25^3\uparrow$ \\
    \toprule

    \multirow{9}{*}{AdaBins}
                            & CL              & 0.222$\pm$0.001 & 1.477$\pm$0.030 & 3.552$\pm$0.008 & 0.254$\pm$0.000 & 0.733$\pm$0.002 & 0.887$\pm$0.001 & 0.941$\pm$0.000  \\
    \cmidrule{2-9}
                            & FedAvg          & \underline{0.434$\pm$0.003} & \underline{2.990$\pm$0.041} & 6.959$\pm$0.070 & 0.542$\pm$0.004 & 0.369$\pm$0.005 & 0.580$\pm$0.004 & 0.729$\pm$0.003  \\
                            & FedProx         & 0.439$\pm$0.003 & 3.029$\pm$0.031 & \underline{6.933$\pm$0.038} & 0.533$\pm$0.002 & 0.369$\pm$0.002 & 0.585$\pm$0.001 & 0.737$\pm$0.001  \\
                            & FedDyn          & 0.440$\pm$0.002 & 3.063$\pm$0.030 & 7.027$\pm$0.073 & 0.551$\pm$0.004 & 0.367$\pm$0.005 & 0.573$\pm$0.004 & 0.721$\pm$0.003  \\
                            & SCAFFOLD        & 0.631$\pm$0.104 & 6.009$\pm$1.035 & 9.664$\pm$0.829 & 0.749$\pm$0.103 & 0.225$\pm$0.036 & 0.432$\pm$0.065 & 0.603$\pm$0.077  \\
    \cmidrule{2-9}
                            & FedClust        & 0.444$\pm$0.004 & 3.263$\pm$0.094 & 7.521$\pm$0.123 & 0.566$\pm$0.014 & 0.277$\pm$0.013 & 0.550$\pm$0.014 & 0.733$\pm$0.012  \\
                            & PACFL$^\dagger$ & 0.602$\pm$0.021 & 6.447$\pm$0.592 & 7.348$\pm$0.142 & \underline{0.509$\pm$0.006} & \underline{0.372$\pm$0.003} & \underline{0.605$\pm$0.004} & \underline{0.768$\pm$0.006}  \\
                            &  Ours           & \textbf{0.318$\pm$0.003} & \textbf{2.127$\pm$0.045} & \textbf{5.354$\pm$0.046} & \textbf{0.364$\pm$0.004} & \textbf{0.515$\pm$0.004} & \textbf{0.770$\pm$0.005} & \textbf{0.897$\pm$0.003}  \\

    \midrule\midrule
    
    \multirow{9}{*}{NeWCRFs}
                            & CL              & 0.166$\pm$0.001 & 0.994$\pm$0.012 & 2.987$\pm$0.009 & 0.190$\pm$0.000 & 0.827$\pm$0.001 & 0.935$\pm$0.000 & 0.968$\pm$0.000  \\
    \cmidrule{2-9}
                            & FedAvg          & 0.366$\pm$0.002 & 1.994$\pm$0.031 & 4.475$\pm$0.020 & 0.406$\pm$0.002 & 0.504$\pm$0.002 & 0.717$\pm$0.001 & 0.834$\pm$0.001  \\
                            & FedProx         & 0.364$\pm$0.003 & 1.983$\pm$0.034 & 4.479$\pm$0.022 & 0.405$\pm$0.002 & 0.504$\pm$0.003 & 0.718$\pm$0.001 & 0.834$\pm$0.001  \\
                            & FedDyn          & 0.363$\pm$0.003 & 1.957$\pm$0.033 & \underline{4.473$\pm$0.022} & 0.404$\pm$0.002 & 0.505$\pm$0.002 & 0.719$\pm$0.001 & 0.834$\pm$0.001  \\
                            & SCAFFOLD        & 0.356$\pm$0.006 & 2.060$\pm$0.125 & 4.569$\pm$0.043 & 0.376$\pm$0.011 & 0.498$\pm$0.024 & 0.726$\pm$0.012 & 0.864$\pm$0.003  \\
    \cmidrule{2-9}
                            & FedClust        & 0.487$\pm$0.003 & 4.325$\pm$0.053 & 7.760$\pm$0.045 & 0.615$\pm$0.007 & 0.241$\pm$0.003 & 0.507$\pm$0.006 & 0.701$\pm$0.006  \\
                            & PACFL$^\dagger$ & \underline{0.318$\pm$0.002} & \underline{1.934$\pm$0.022} & 4.265$\pm$0.018 & \underline{0.327$\pm$0.002} & \underline{0.540$\pm$0.002} & \underline{0.791$\pm$0.002} & \underline{0.928$\pm$0.002}  \\
                            & Ours            & \textbf{0.249$\pm$0.001} & \textbf{1.750$\pm$0.014} & \textbf{3.838$\pm$0.011} & \textbf{0.261$\pm$0.001} & \textbf{0.703$\pm$0.002} & \textbf{0.892$\pm$0.001} & \textbf{0.954$\pm$0.000}  \\
    \midrule\midrule

    \multirow{9}{*}{DCDepth}
                            &    CL           & 0.159$\pm$0.000 & 0.962$\pm$0.010 & 2.916$\pm$0.004 & 0.183$\pm$0.000 & 0.835$\pm$0.000 & 0.939$\pm$0.000 & 0.971$\pm$0.000  \\
    \cmidrule{2-9}
                            & FedAvg          & 0.351$\pm$0.004 & 2.077$\pm$0.089 & 4.640$\pm$0.041 & 0.348$\pm$0.002 & 0.493$\pm$0.006 & 0.766$\pm$0.001 & 0.902$\pm$0.000  \\
                            & FedProx         & 0.354$\pm$0.007 & 2.152$\pm$0.083 & 4.641$\pm$0.053 & 0.348$\pm$0.004 & 0.492$\pm$0.005 & 0.769$\pm$0.004 & 0.902$\pm$0.002  \\
                            & FedDyn          & 0.352$\pm$0.003 & 2.109$\pm$0.034 & 4.699$\pm$0.043 & 0.349$\pm$0.002 & 0.491$\pm$0.006 & 0.767$\pm$0.001 & 0.903$\pm$0.000  \\
                            & SCAFFOLD        & 0.329$\pm$0.005 & 2.050$\pm$0.055 & 4.880$\pm$0.104 & 0.350$\pm$0.004 & 0.472$\pm$0.014 & 0.759$\pm$0.005 & 0.914$\pm$0.003  \\
    \cmidrule{2-9}
                            & FedClust        & 0.349$\pm$0.045 & 3.132$\pm$0.753 & 6.423$\pm$0.551 & 0.434$\pm$0.039 & 0.498$\pm$0.046 & 0.712$\pm$0.047 & 0.815$\pm$0.045  \\
                            & PACFL$^\dagger$ & \underline{0.321$\pm$0.002} & \underline{1.976$\pm$0.028} & \textbf{4.338$\pm$0.017} & \underline{0.325$\pm$0.001} & \underline{0.537$\pm$0.001} & \underline{0.796$\pm$0.001} & \underline{0.924$\pm$0.001}  \\
                            & Ours            & \textbf{0.293$\pm$0.001} & \textbf{1.872$\pm$0.015} & \underline{4.469$\pm$0.024} & \textbf{0.310$\pm$0.001} & \textbf{0.547$\pm$0.006} & \textbf{0.840$\pm$0.001} & \textbf{0.941$\pm$0.001}  \\
    
    \bottomrule
  \end{tabular}
}
  } \normalsize

  \label{tab:hpe_full_results}
\end{minipage}
}
\end{table}

\begin{table}
\centering
\rotatebox{90}{
\begin{minipage}{0.98\textheight}
  \caption{
  {\textbf{Quantitative evaluation on the BMR scenario.}}
  } 
\resizebox{\linewidth}{!}{%
\setlength{\tabcolsep}{5pt}
  \centering
  { \small
    \begin{tabular}{c|>{\centering\arraybackslash}m{2cm}|cccc|ccc}
    \toprule
    \textbf{Baseline} & \textbf{Method} & $Abs\ Rel\downarrow$ & $Sq\ Rel\downarrow$ & $RMSE\downarrow$ & $RMSE_{\log}\downarrow$ & $\delta < 1.25\uparrow$ & $\delta < 1.25^2\uparrow$ & $\delta < 1.25^3\uparrow$ \\
    \toprule

    \multirow{9}{*}{AdaBins}
                            & CL              & 0.166$\pm$0.054 & 0.369$\pm$0.320 & 2.073$\pm$0.613 & 0.191$\pm$0.049 & 0.768$\pm$0.084 & 0.945$\pm$0.059 & 0.984$\pm$0.033  \\
    \cmidrule{2-9}
                            & FedAvg          & 0.357$\pm$0.003 & 1.153$\pm$0.017 & 3.329$\pm$0.024 & 0.409$\pm$0.004 & 0.211$\pm$0.005 & 0.673$\pm$0.007 & 0.913$\pm$0.005  \\
                            & FedProx         & 0.381$\pm$0.001 & 1.316$\pm$0.014 & 3.439$\pm$0.017 & 0.430$\pm$0.002 & 0.193$\pm$0.001 & 0.630$\pm$0.005 & 0.899$\pm$0.004  \\
                            & FedDyn          & 0.359$\pm$0.002 & 1.160$\pm$0.017 & 3.331$\pm$0.021 & 0.412$\pm$0.003 & 0.212$\pm$0.002 & 0.667$\pm$0.006 & 0.909$\pm$0.005  \\
                            & SCAFFOLD        & 0.424$\pm$0.037 & 2.343$\pm$0.466 & 5.447$\pm$0.086 & 0.501$\pm$0.003 & 0.276$\pm$0.004 & 0.561$\pm$0.006 & 0.804$\pm$0.015  \\
    \cmidrule{2-9}
                            & FedClust        & 0.133$\pm$0.001 & 0.316$\pm$0.006 & 2.220$\pm$0.031 & 0.175$\pm$0.001 & 0.813$\pm$0.004 & 0.964$\pm$0.001 & \underline{0.993$\pm$0.000}  \\
                            & PACFL$^\dagger$ & \textbf{0.124$\pm$0.001} & \textbf{0.263$\pm$0.002} & \textbf{2.022$\pm$0.019} & \textbf{0.164$\pm$0.001} & \textbf{0.831$\pm$0.002} & \textbf{0.970$\pm$0.001} & \textbf{0.994$\pm$0.000} \\ 
                            &  Ours           & \underline{0.126$\pm$0.001} & \underline{0.283$\pm$0.004} & \underline{2.150$\pm$0.024} & \underline{0.167$\pm$0.001} & \underline{0.827$\pm$0.002} & \underline{0.968$\pm$0.001} & \textbf{0.994$\pm$0.000} \\ 

    \midrule\midrule
    
    \multirow{9}{*}{NeWCRFs}
                            & CL              & 0.077$\pm$0.001 & 0.100$\pm$0.000 & 1.209$\pm$0.005 & 0.100$\pm$0.001 & 0.950$\pm$0.001 & 0.994$\pm$0.000 & 0.998$\pm$0.000  \\
    \cmidrule{2-9}
                            & FedAvg          & 0.323$\pm$0.003 & 1.048$\pm$0.020 & 3.094$\pm$0.026 & 0.344$\pm$0.003 & 0.191$\pm$0.003 & 0.833$\pm$0.007 & 0.983$\pm$0.001  \\
                            & FedProx         & 0.330$\pm$0.002 & 1.084$\pm$0.017 & 3.141$\pm$0.022 & 0.350$\pm$0.003 & 0.181$\pm$0.003 & 0.819$\pm$0.007 & 0.981$\pm$0.001  \\
                            & FedDyn          & 0.330$\pm$0.003 & 1.081$\pm$0.021 & 3.136$\pm$0.027 & 0.351$\pm$0.003 & 0.182$\pm$0.004 & 0.816$\pm$0.007 & 0.980$\pm$0.001  \\
                            & SCAFFOLD        & 0.315$\pm$0.009 & 0.836$\pm$0.112 & 2.720$\pm$0.145 & 0.357$\pm$0.007 & 0.292$\pm$0.065 & 0.742$\pm$0.054 & 0.961$\pm$0.014  \\
    \cmidrule{2-9}
                            & FedClust        & 0.105$\pm$0.001 & \underline{0.224$\pm$0.004} & 1.716$\pm$0.010 & 0.135$\pm$0.001 & \underline{0.890$\pm$0.002} & \underline{0.984$\pm$0.000} & \underline{0.997$\pm$0.000}  \\
                            & PACFL$^\dagger$ & \textbf{0.087$\pm$0.001} & \textbf{0.145$\pm$0.002} & \underline{1.501$\pm$0.009} & \textbf{0.117$\pm$0.001} & \textbf{0.921$\pm$0.002} & \textbf{0.991$\pm$0.000} & \textbf{0.999$\pm$0.000}  \\
                            & Ours            & \underline{0.089$\pm$0.001} & \textbf{0.145$\pm$0.002} & \textbf{1.499$\pm$0.009} & \underline{0.118$\pm$0.001} & \textbf{0.921$\pm$0.002} & \textbf{0.991$\pm$0.000} & \textbf{0.999$\pm$0.000}  \\
    \midrule\midrule

    \multirow{9}{*}{DCDepth}
                            &    CL           & 0.074$\pm$0.001 & 0.101$\pm$0.003 & 1.246$\pm$0.027 & 0.099$\pm$0.001 & 0.952$\pm$0.001 & 0.994$\pm$0.000 & 0.998$\pm$0.000  \\
    \cmidrule{2-9}
                            & FedAvg          & 0.158$\pm$0.001 & 0.265$\pm$0.002 & 1.789$\pm$0.009 & 0.183$\pm$0.001 & 0.801$\pm$0.002 & 0.969$\pm$0.000 & \underline{0.996$\pm$0.000}  \\
                            & FedProx         & 0.168$\pm$0.000 & 0.303$\pm$0.004 & 1.937$\pm$0.014 & 0.193$\pm$0.001 & 0.771$\pm$0.003 & 0.963$\pm$0.000 & 0.994$\pm$0.000  \\
                            & FedDyn          & 0.171$\pm$0.001 & 0.299$\pm$0.005 & 1.880$\pm$0.025 & 0.197$\pm$0.001 & 0.751$\pm$0.003 & 0.963$\pm$0.000 & 0.994$\pm$0.000  \\
                            & SCAFFOLD        & 0.209$\pm$0.004 & 0.518$\pm$0.141 & 2.602$\pm$0.571 & 0.247$\pm$0.022 & 0.599$\pm$0.067 & 0.916$\pm$0.029 & 0.986$\pm$0.004  \\
    \cmidrule{2-9}
                            & FedClust        & \underline{0.087$\pm$0.001} & \underline{0.141$\pm$0.001} & 1.437$\pm$0.005 & \underline{0.112$\pm$0.001} & \underline{0.929$\pm$0.001} & \underline{0.992$\pm$0.000} & \textbf{0.999$\pm$0.000}  \\
                            & PACFL$^\dagger$ & \textbf{0.082$\pm$0.000} & \textbf{0.130$\pm$0.001} & \textbf{1.428$\pm$0.006} & \underline{0.109$\pm$0.001} & \textbf{0.933$\pm$0.001} & \textbf{0.993$\pm$0.000} & \textbf{0.999$\pm$0.000}  \\
                            & Ours            & \textbf{0.082$\pm$0.000} & \textbf{0.127$\pm$0.001} & \underline{1.430$\pm$0.003} & \textbf{0.108$\pm$0.001} & \textbf{0.933$\pm$0.001} & \textbf{0.993$\pm$0.000} & \textbf{0.999$\pm$0.000}  \\
    
    \bottomrule
  \end{tabular}
}
  } \normalsize

  \label{tab:bmr_full_results}
\end{minipage}
}
\end{table}

\section{Additional Experiments}
\label{sec:additional_experiments}

\noindent \textbf{Additional quantitative evaluations.} \
Tabs.~\ref{tab:hpe_full_results} and \ref{tab:bmr_full_results} present the complete quantitative evaluations for all compared methods under the HPE and BMR scenarios, respectively. 
%
%
While the main paper reports the mean performance of representative methods over the last five communication rounds for readability, this supplementary material provides the full results along with standard deviations for a comprehensive comparison.
These results complement the main paper by revealing not only the average performance but also the standard deviations and convergence stability of each method across communication rounds.

As shown in Tab.~\ref{tab:hpe_full_results}, \textsc{FeDepth} consistently achieves competitive performance under the HPE setting despite the presence of diverse robot platforms and environmental conditions.
In addition, the reported standard deviations and the convergence graph (Fig.~\ref{fig:graph_for_HPE}) indicate that ~\textsc{FeDepth} exhibits the most stable convergence among the compared methods.

Tab.~\ref{tab:bmr_full_results} presents the evaluations under the BMR setting, which represents an extremely heterogeneous non-IID condition composed of distinct indoor and outdoor domains. 
%
Conventional FL optimization methods struggle under such severe heterogeneity.
In particular, not only FedAvg~\cite{fedavg} but also stabilization-based approaches such as FedProx~\cite{li2020fedprox}, FedDyn~\cite{acar2021feddyn}, and SCAFFOLD~\cite{karimireddy2020scaffold} fail to consistently improve performance. 
%
These results suggest that learning a single global model that generalizes well across clients is challenging when the underlying data distributions are highly separated.

\begin{figure*}[!t]
    \centering
    \includegraphics[width=\textwidth]{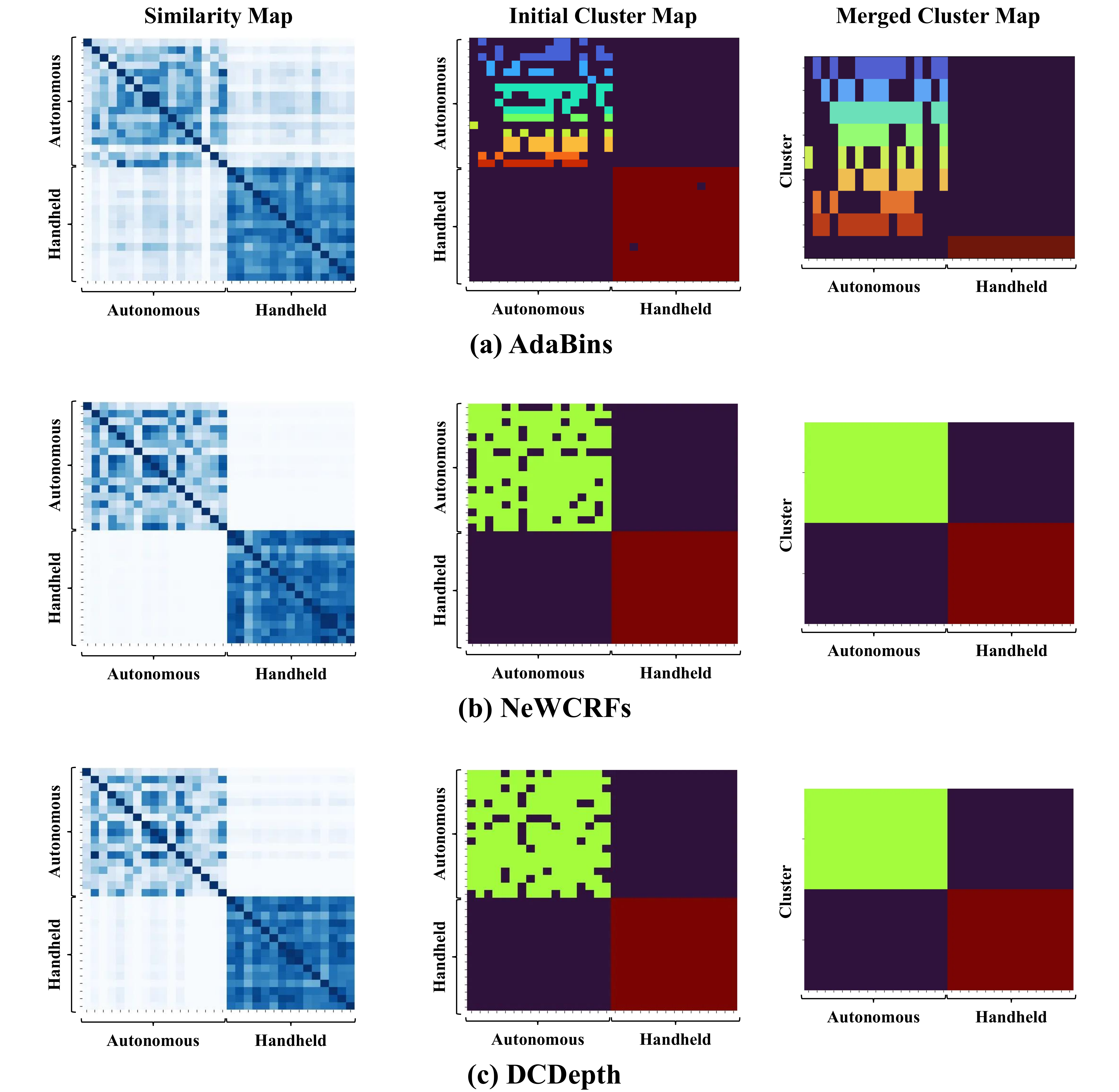}
    \caption{\textbf{Visualization of the clustering result on the BMR scenario.}}
    \label{fig:bmr_cluster}
\end{figure*}

In contrast, CFL-based approaches that explicitly separate models via clustering are more effective under such severe heterogeneity.
%
Notably, when domain boundaries are clear, \textsc{FeDepth} naturally produces nearly hard-cluster assignments, as illustrated in Fig.~\ref{fig:bmr_cluster}.
Consequently, the performance gap between \textsc{FeDepth} and the hard-clustering baseline PACFL$^{\dagger}$~\cite{vahidian2023efficientpacfl} is effectively reduced.

\begin{figure*}[t]
    \centering
    \includegraphics[width=0.99\textwidth]{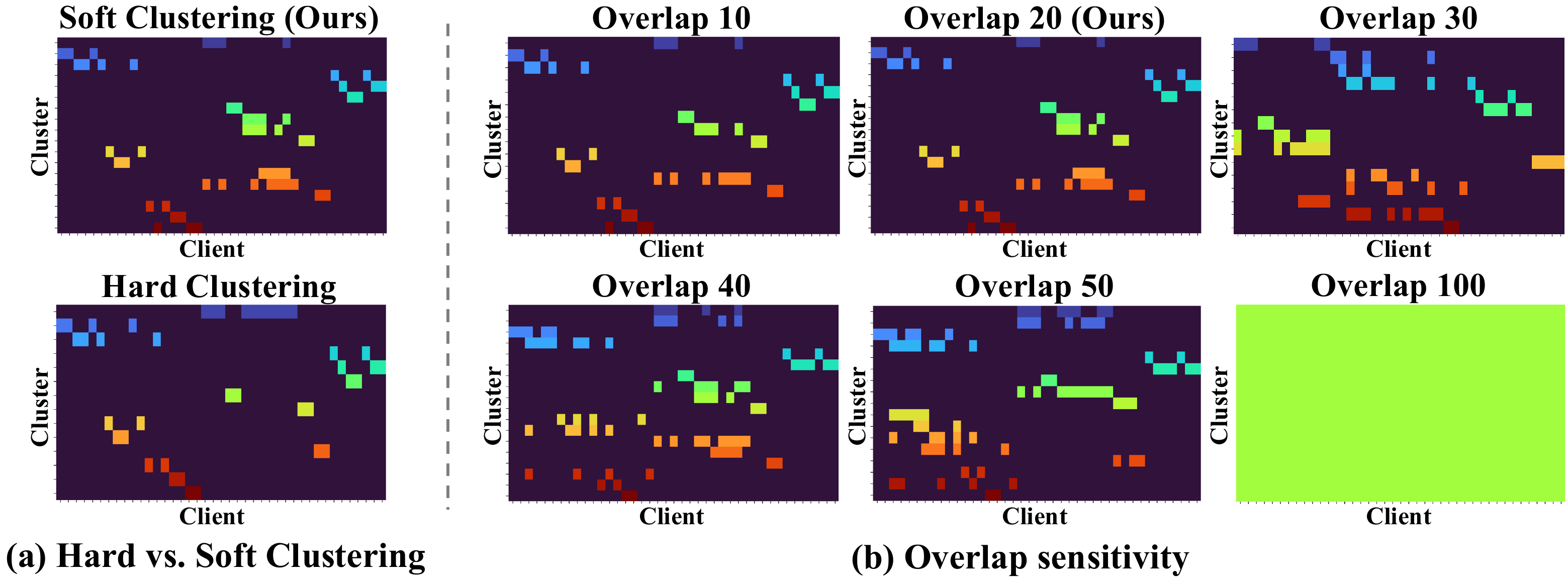}
    \caption{\textbf{Visualization of the clustering results for the ablation study.}}
    \label{fig:additional_ablation_viz}
\end{figure*}

\vspace{1mm}
\vspace{1mm} \noindent \textbf{Additional results for the ablation study.} \ 
To provide further insight into the ablation results reported in Tab.~3 of the main paper, we visualize the resulting cluster structures in Fig.~\ref{fig:additional_ablation_viz}.
Fig.~\ref{fig:additional_ablation_viz}a compares the cluster assignments obtained by hard and soft clustering under the same threshold $\tau$. 
%
Hard clustering assigns each client to a single cluster, which can be restrictive for clients near cluster boundaries.
%
In contrast, soft clustering allows clients to belong to multiple clusters, enabling more flexible associations across related domains.
%
Fig.~\ref{fig:additional_ablation_viz}b shows the clustering results obtained with different overlap ratios.
As the overlap increases, more clients participate in multiple clusters, whereas an overlap of $100\%$ collapses all clients into a single cluster, making the setting equivalent to FedAvg~\cite{fedavg}. 
%
Consistent with the quantitative results in Tab.~3b, an overlap ratio of $20\%$ yields the most balanced clustering result.

\begin{table}[t]
\centering
\caption{\textbf{Comparison with personalized federated learning methods.} All results are based on NeWCRFs~\cite{newcrfs} under the HPE scenario.}
\vspace{-3mm}
\label{tab:pfl_compare}
\resizebox{0.95\linewidth}{!}{%
\setlength{\tabcolsep}{5pt}
    \centering
    { \small
    \begin{tabular}{c|cccc|ccc}
        \toprule
        Method & \textit{Abs Rel} $\downarrow$ & \textit{Sq Rel} $\downarrow$ & \textit{RMSE} $\downarrow$ & \textit{RMSE}$_{\log}$ $\downarrow$ & $\delta < 1.25$ $\uparrow$ & $\delta < 1.25^2$ $\uparrow$ & $\delta < 1.25^3$ $\uparrow$ \\
        \midrule
        FedAvg & \underline{0.366} & \underline{1.994} & 4.475 & \underline{0.406} & \underline{0.504} & \underline{0.717} & \underline{0.834} \\
        pFedMe & 0.563 & 3.978 & \underline{4.227} & 0.446 & 0.477 & 0.710 & 0.800 \\
        \midrule
        Ours & \textbf{0.249} & \textbf{1.750} & \textbf{3.838} & \textbf{0.261} & \textbf{0.703} & \textbf{0.892} & \textbf{0.954} \\
        \bottomrule
    \end{tabular}
    }
  }\normalsize
  \vspace{-4mm}
\end{table}
\vspace{1mm} \noindent \textbf{Comparison with personalized federated learning.} \ 
Tab.~\ref{tab:pfl_compare} compares \textsc{FeDepth} with a personalized federated learning (pFL) method, pFedMe~\cite{pfedme}.
pFL methods typically rely on client-specific data to construct or adapt personalized models for each client.
%
In contrast, our robotic setting focuses on generalization to unseen environments, where the test sequence comes from a previously unseen client and such data is unavailable at inference time.
%
Under this setting, the standard personalization procedure used in pFL methods has limited applicability.
%
Therefore, to ensure a fair comparison, we evaluate the global model of pFedMe rather than its personalized models.
%
Although this setup limits the performance of pFL methods, it reflects the misalignment between their personalization paradigm and our robotic setting.
%
In contrast, \textsc{FeDepth} selects a suitable cluster model based on descriptor similarity in a zero-shot manner, which is better aligned with generalization to unseen domains.

\begin{figure*}[!t]
    \centering
    \includegraphics[width=\textwidth]{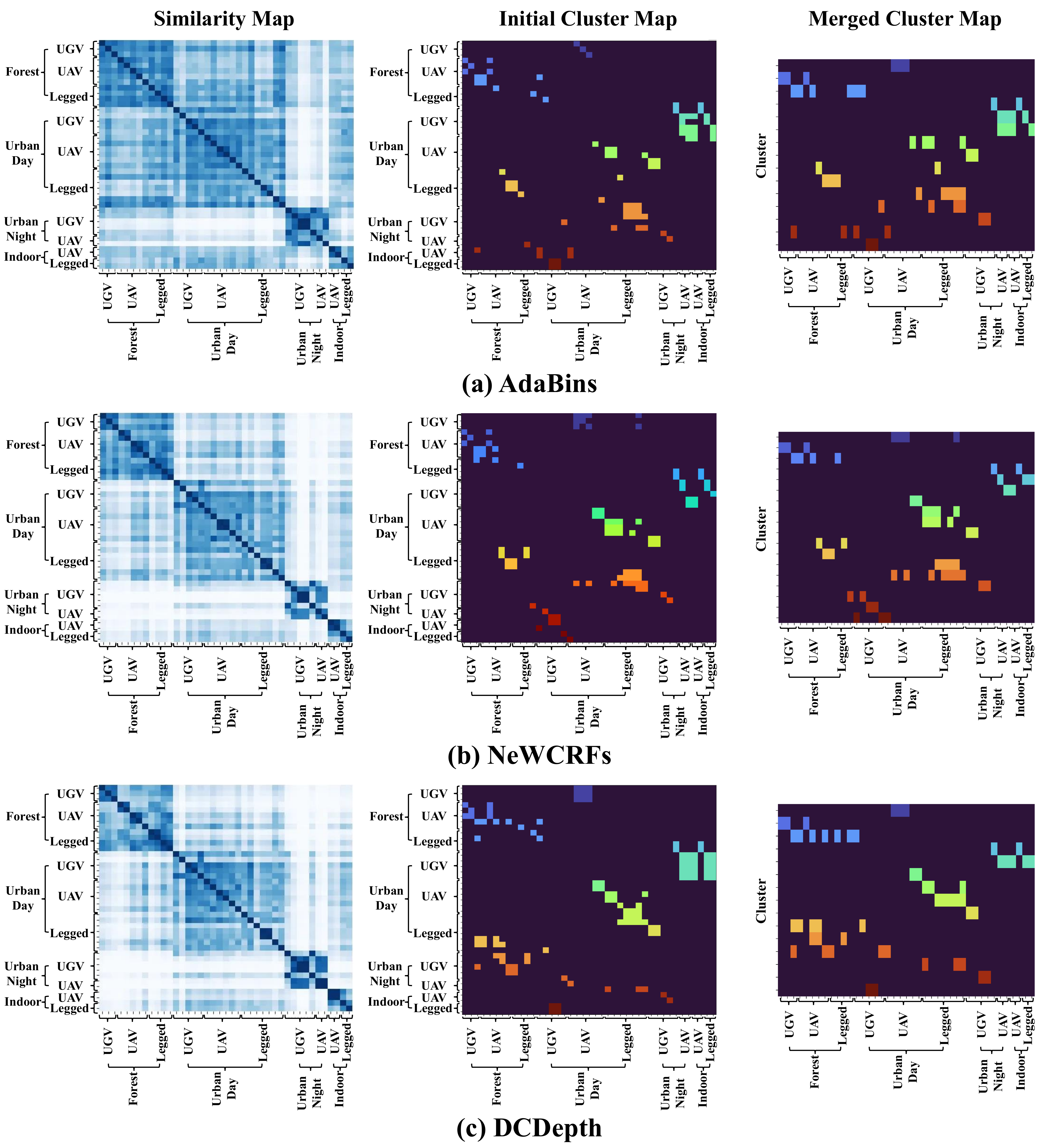}
    \caption{\textbf{Visualization of the clustering result on the HPE scenario.}}
    \label{fig:hpe_cluster}
\end{figure*}


\section{Clustering Results}
\label{sec:clustering_result}

Figs.~\ref{fig:hpe_cluster} and \ref{fig:bmr_cluster} show the clustering behavior of our method across two heterogeneous scenarios: HPE and BMR. In the cluster map, the same color denotes the same cluster.

\vspace{1mm} \noindent \textbf{HPE Scenario.} \ 
This setting involves diverse platform types (UGV, UAV, Legged) and environmental conditions (forest, urban-day, urban-night, indoor), creating complex domain boundaries. 
%
Consequently, regardless of the encoder used, initial clustering based on a simple threshold fails to yield a structured representation of client distributions, often resulting in fragmented and disorganized groups.
For instance, we observe redundant clusters that are merely subsets of others, or even isolated singleton groups, which do not effectively capture the global data structure.
In response, $\textsc{FeDepth}$ effectively consolidates these clusters through merging, refining the structure so that the resulting groups more coherently represent the client population.
Moreover, by permitting clients to participate in multiple groups, our framework captures these complex inter-client relationships with greater fidelity.



\vspace{1mm} \noindent \textbf{BMR Scenario.} \ 
This setting consists of two distinct domains: autonomous driving and handheld capture. Consequently, the similarity matrices across all baselines display two clearly separated distributions. %
Ideally, clustering in this scenario should yield exactly two groups corresponding to the distinct datasets. Therefore, hard clustering methods such as PACFL$^\dagger$ are well-suited to represent this global data distribution, leading to the performance gains observed in Tab.~\ref{tab:bmr_full_results}. 
For $\textsc{FeDepth}$, the distance distribution also exhibits a clear separation between KITTI (autonomous) and NYUv2 (handheld). However, due to variations in inter-client distances even within the same dataset (see the first column of Fig.~\ref{fig:bmr_cluster}), the initial clustering does not always result in a perfect two-cluster separation. 
Similar to the HPE scenario, our cluster merging process r

efines these initial groups to better represent the global data structure. Consequently, for NeWCRFs and DCDepth, the method successfully recovers the ideal two-cluster configuration (see Figs.~\ref{fig:bmr_cluster}b and c). 
In the case of AdaBins (see Fig.~\ref{fig:bmr_cluster}a), however, the encoder measures large distances even between clients within the same dataset, which hinders the formation of ideal clusters. 
Nevertheless, as shown in Tab.~\ref{tab:bmr_full_results}, \textsc{FeDepth} achieves performance comparable to specialized hard clustering methods, demonstrating its robustness. 
%
In conclusion, these results confirm that \textsc{FeDepth} operates effectively in both complex, entangled scenarios (HPE) and settings theoretically less favorable for soft clustering (BMR).

\section{Additional Qualitative Results}
\label{sec:visualization_result}



\begin{figure}[H]
    \vspace{-4mm}
    \centering
    \includegraphics[width=0.99\textwidth]{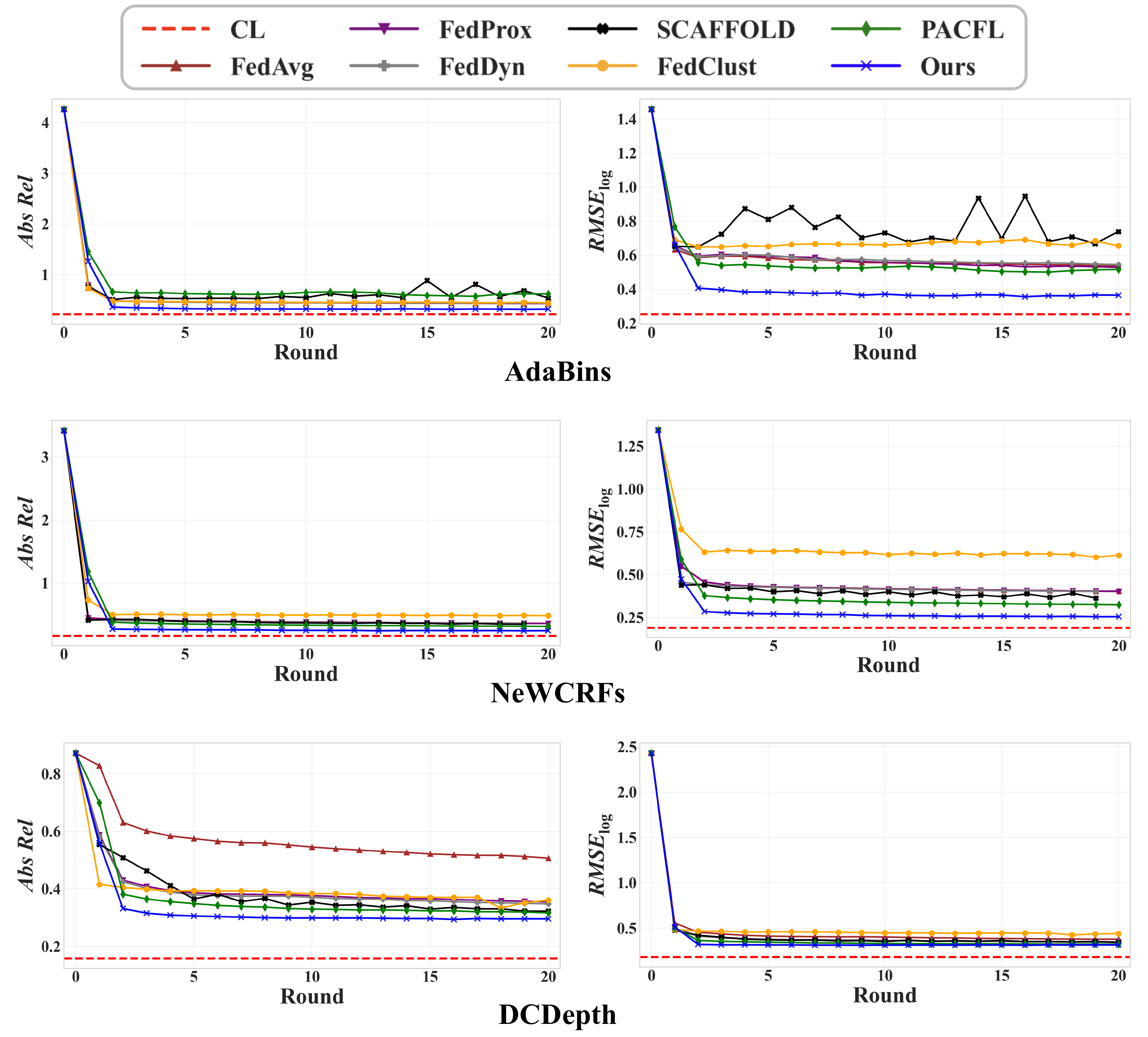}
    \caption{\textbf{{Convergence graph on the HPE scenario}.}}
    \label{fig:graph_for_HPE}
\end{figure}

\begin{figure}[ht]
    \vspace{-4mm}
    \centering
    \includegraphics[width=0.99\textwidth]{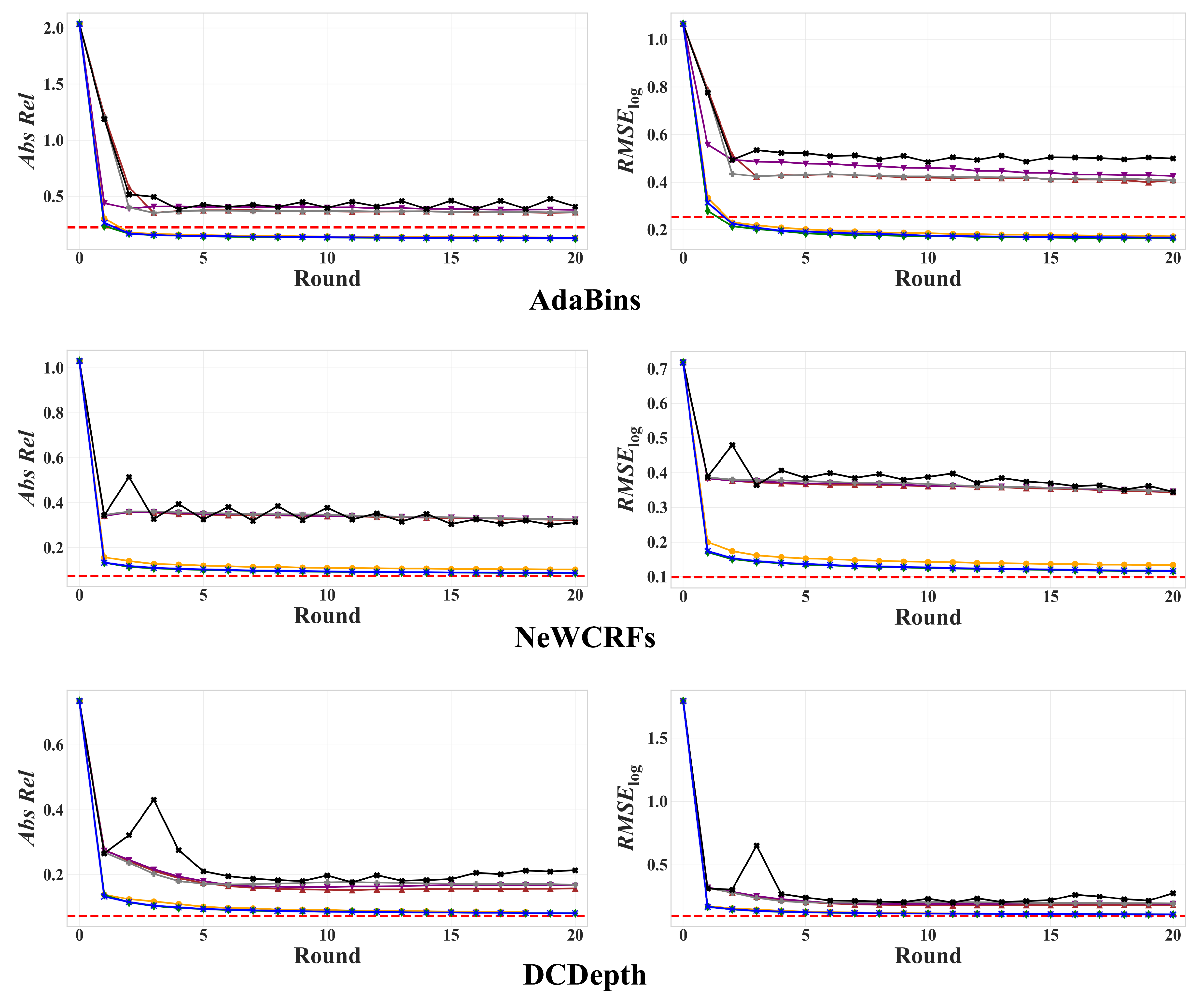}
    \caption{\textbf{{Convergence graph on the BMR scenario}.}}
    \label{fig:graph_for_BMR}
\end{figure}

\noindent \textbf{Convergence graphs.} \ 
Convergence graphs of the Abs Rel and $RMSE_{\log}$ errors over communication rounds are presented for both scenarios. 
For the HPE scenario (see Fig.~\ref{fig:graph_for_HPE}), \textsc{FeDepth} exhibits fast and stable convergence throughout training. 
Across all depth estimation baselines in the BMR scenario (see Fig.~\ref{fig:graph_for_BMR}), \textsc{FeDepth} exhibits convergence comparable to CL, highlighting the effectiveness of clustering under significant domain gaps.

\begin{figure*} 
\rotatebox{90}{
\resizebox{0.97\textheight}{!}{
\begin{minipage}{\textheight}
    \centering
    \includegraphics[width=\textheight]{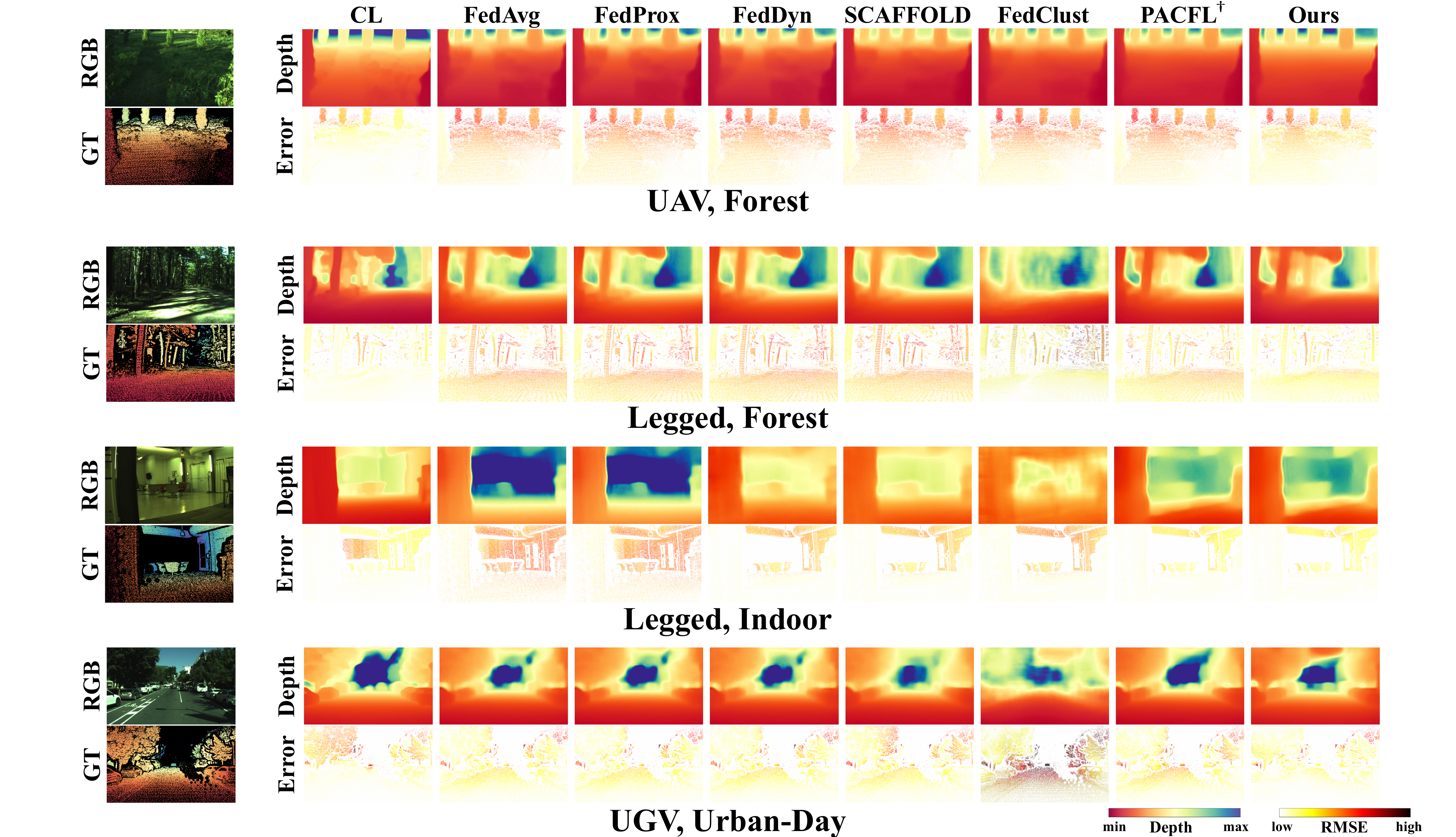}
    \caption{\textbf{Qualitative comparison of different FL methods on the HPE scenario using DCDepth}~\cite{dcdepth}. The first column shows the RGB image and ground-truth depth, while each method presents the predicted depth (top) and error map (bottom). Labels (\emph{e.g.}, UAV, Forest) indicate the platform and environment.}
    \label{fig:hpe_depth}
\end{minipage}
}}
\end{figure*}

\begin{figure*} 
    \centering
    \includegraphics[width=0.98\textwidth]{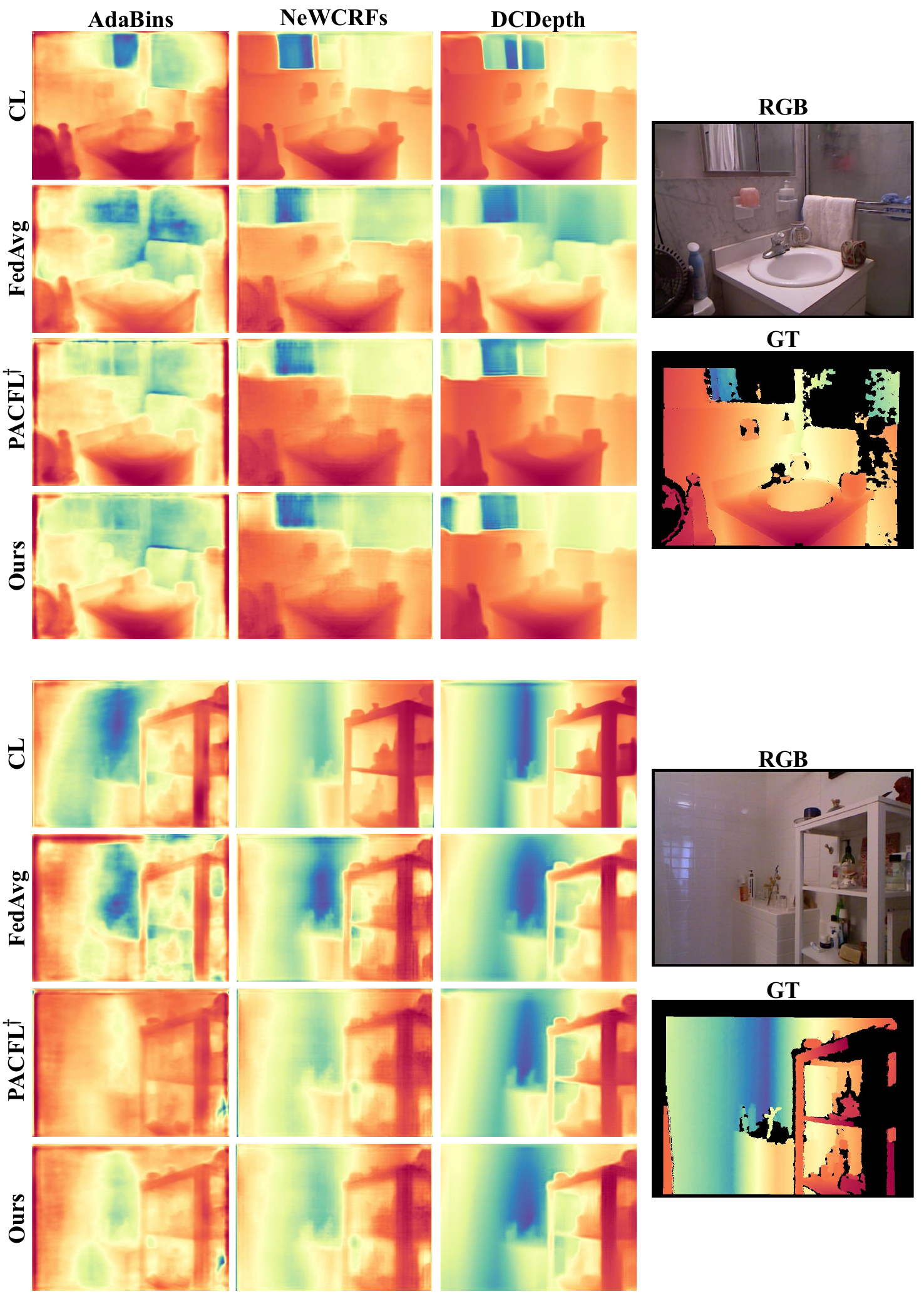}
    \caption{\textbf{Qualitative comparison of different depth estimation baselines on the BMR scenario (Handheld)}.}
    \label{fig:bmt_depth_NYU}
\end{figure*}

\begin{figure*} 
    \centering
    \includegraphics[width=0.98\textwidth]{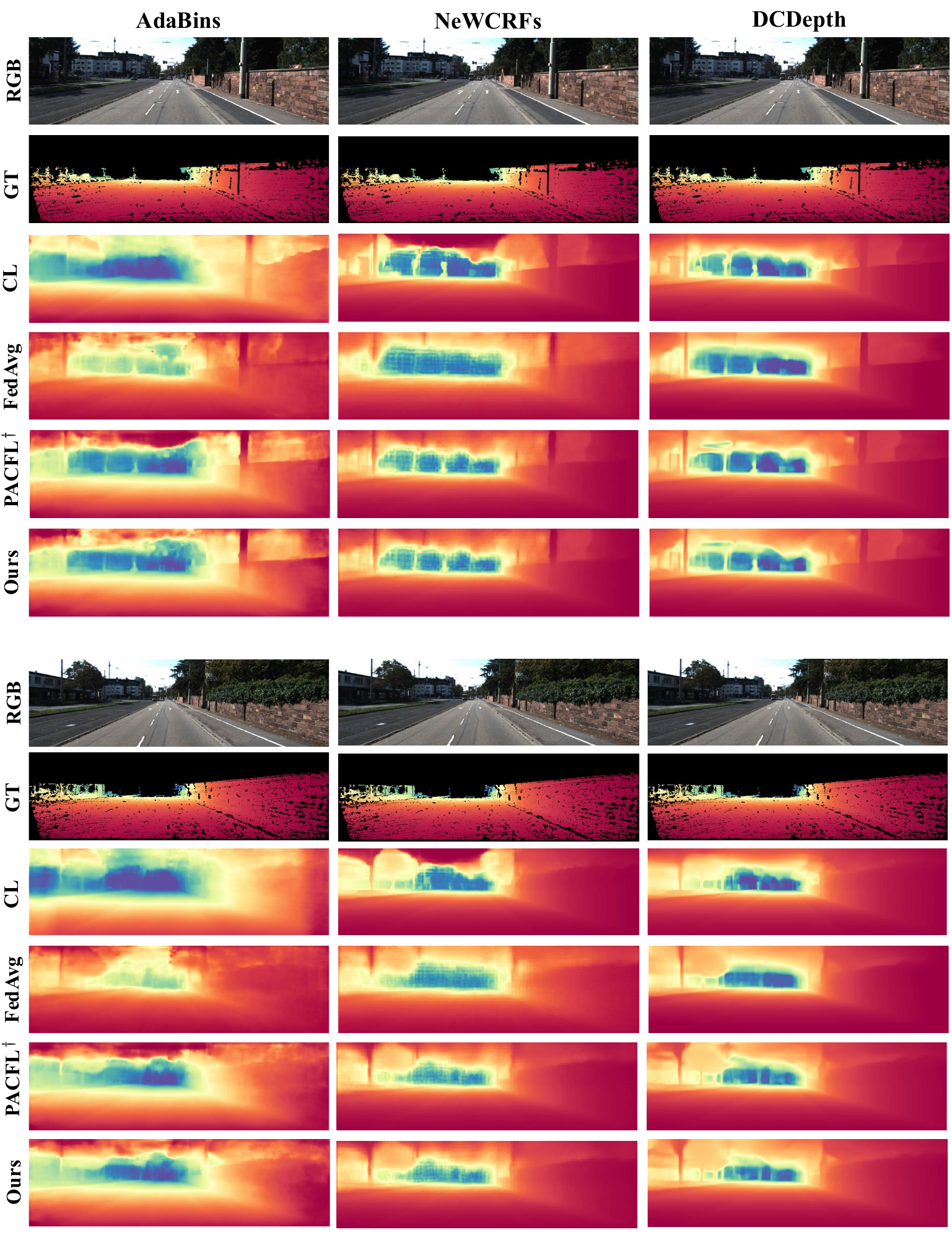}
    \caption{\textbf{Qualitative comparison of different depth estimation baselines on the BMR scenario (Autonomous)}.}
    \label{fig:bmt_depth_KITTI}
\end{figure*}

\vspace{1mm} \noindent \textbf{Qualitative depth results.} \ 
We provide qualitative results for the HPE scenario (Fig.~\ref{fig:hpe_depth}) and BMR scenario (Figs.~\ref{fig:bmt_depth_NYU} and \ref{fig:bmt_depth_KITTI}).
In Fig.~\ref{fig:hpe_depth}, \textsc{FeDepth} delivers more accurate and refined depth estimates than standard FL in the HPE setting.
Furthermore, error maps show that \textsc{FeDepth} captures depth ranges more precisely than FedClust~\cite{islam2024fedclust} and PACFL$^\dagger$~\cite{vahidian2023efficientpacfl}.
Although the estimated depth maps exhibit similar levels of sharpness across all baselines, the error maps clearly reveal that \textsc{FeDepth} distinguishes depth ranges more effectively.

The BMR scenario is characterized by two distinctly separated data distributions. As discussed in Sec.~\ref{sec:additional_experiments}, hard clustering methods, which partition clients into groups for independent training, are well-suited for such distributions.
Furthermore, we confirm that $\textsc{FeDepth}$, despite utilizing soft clustering, also operates effectively in this setting. 
As illustrated in Figs.~\ref{fig:bmt_depth_NYU} and \ref{fig:bmt_depth_KITTI}, this strategy consistently yields improvements regardless of the depth estimation baseline.

%





\section{Implementation Details}
\label{sec:implementation_detail}

\begin{figure}[H]
    \vspace{-3mm}
    \centering
    \includegraphics[width=0.9\linewidth]{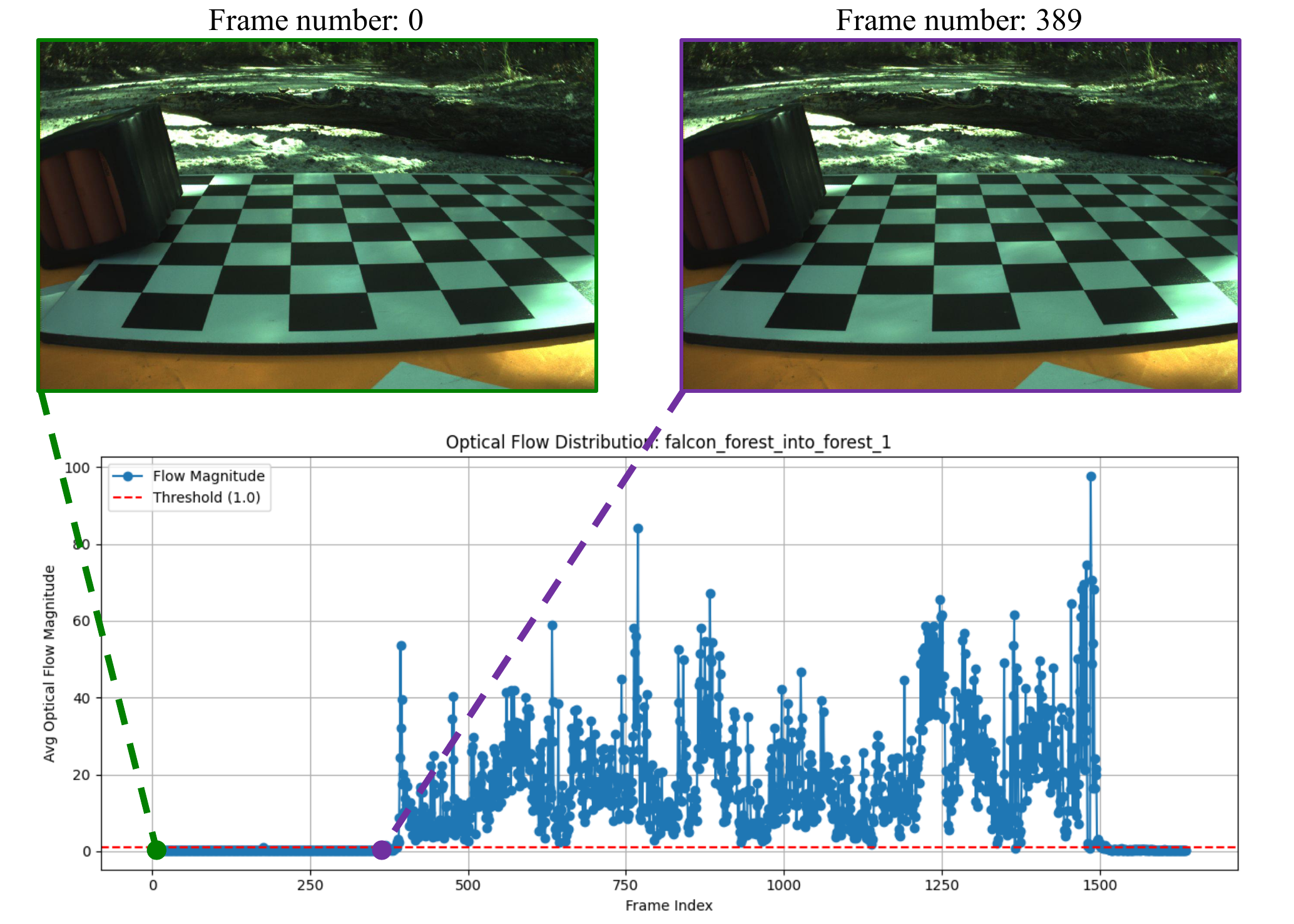}    \caption{\textbf{Optical flow visualization results for sequence falcon forest into forest 1.} Even though there is a difference of more than 10 seconds between the frame marked with a green dotted line (upper left) and the purple frame (upper right), there is no movement. To remove these stationary frames, we remove those where the average optical flow within the image pixels is lower than the threshold (marked with a red dotted line in the histogram).}
    \label{fig:optical}
\end{figure}

\noindent \textbf{Preprocessing.} \
The M3ED dataset~\cite{m3ed} provides raw robot-captured sequences, which may include noise and other signals that are not directly useful for training.
To mitigate this, we first detect and remove static frames (i.e., periods without motion) by computing optical flow~\cite{lucas1981iterative} using the OpenCV library~\cite{bradski2008learning} following prior work~\cite{zhou2017unsupervised}, as illustrated in Fig.~\ref{fig:optical}.
After filtering out these frames, we construct valid ground-truth data for monocular depth estimation. The official ground-truth depth maps provided by M3ED are aligned with event-camera frames rather than RGB camera frames. To obtain depth maps directly aligned with the RGB images, we project LiDAR point clouds onto each RGB frame and extract new ground-truth depth maps.
%
For the HPE scenario, the corresponding depth maps are also downsampled to match the input image resolution.

We use different train/test splits for the BMR and HPE scenarios.
The BMR setting is designed to explicitly evaluate the effectiveness of clustering.
Following prior work~\cite{bhat2023zoedepth}, we evaluate the models on NYUv2 and KITTI, which correspond to clearly separated indoor and outdoor domains, respectively.
Because the data distributions are distinctly separated, clustered federated learning (CFL) can achieve strong performance even with relatively simple clustering strategies.
In contrast, the HPE scenario is designed to reflect conditions closer to real-world robot perception, where domain shifts are more complex and overlapping. 
Accordingly, we curate the test set to cover a wide range of non-overlapping combinations of platforms and observation environments.

Through these two scenarios, we analyze federated learning performance under different domain conditions. 
The BMR scenario evaluates the fundamental impact of clearly separated indoor and outdoor depth distributions, as well as platform and environment differences. 
In contrast, the HPE scenario enables a deeper analysis of how federated learning models behave under complex and overlapping domain variations encountered in realistic robot perception settings.
%

\begin{figure*}[!htbp]
    \centering
    \includegraphics[width=0.85\linewidth]{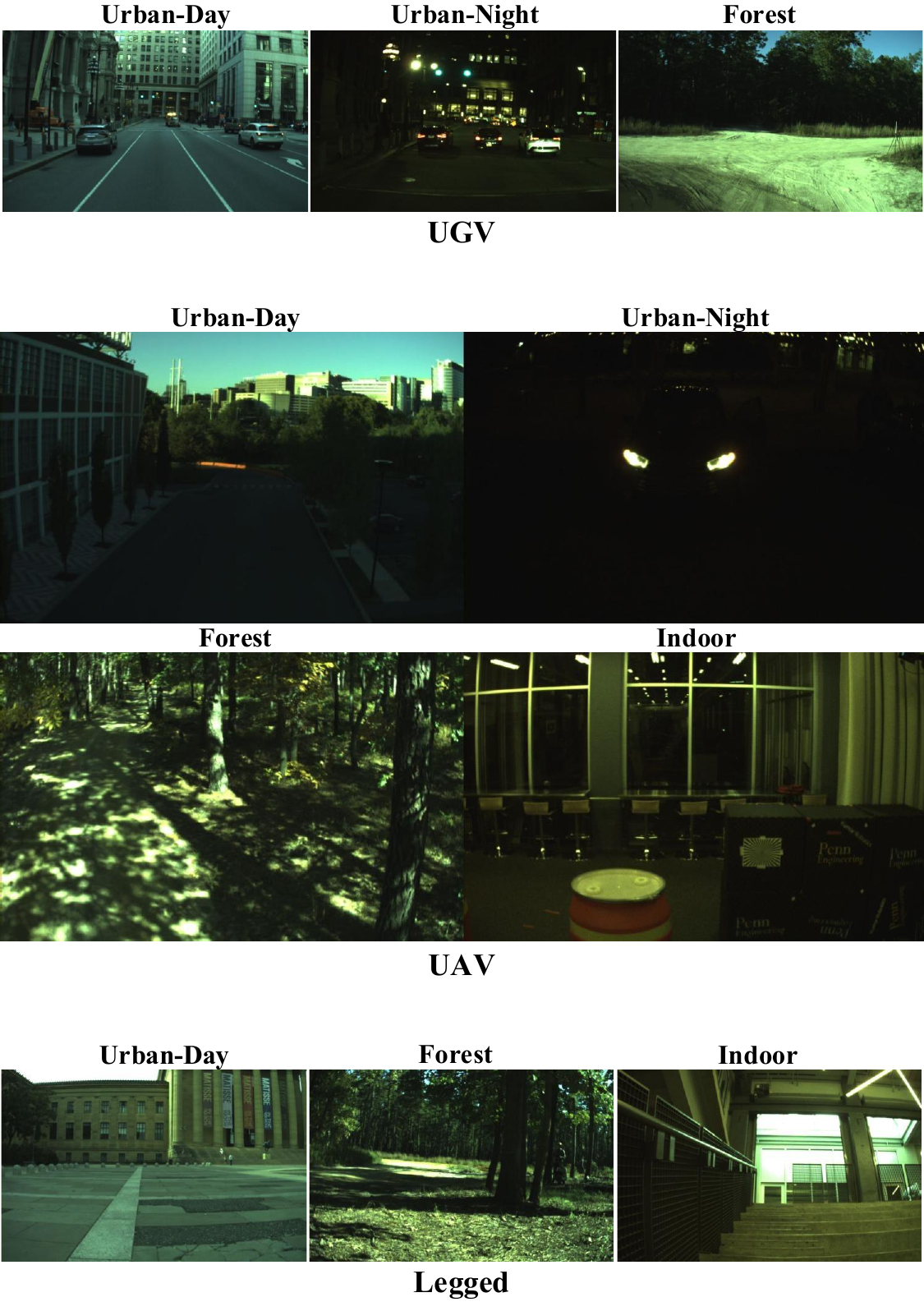}
    \caption{\textbf{Representative RGB images from different platforms and environments used in the HPE scenario.}}
    \label{fig:HPE_visualize}
\end{figure*}

%
\vspace{1mm} \noindent \textbf{Scenario details.} \ 
As mentioned in the main paper, \textsc{FeDepth} redefines the attribute structure of the M3ED~\cite{m3ed} dataset to construct the HPE scenario.
%
As shown in Tab.~\ref{tab:HPE_sequences_sorted}, we categorize environments into four groups, namely indoor, forest, urban day, and urban night, each exhibiting distinct characteristics.
The indoor category covers scenes with depth ranges ($\sim$10 m), typically captured in staircases and indoor rooms, showing planar surfaces and textureless regions.
The forest category consists of unstructured natural scenes featuring repetitive patterns such as trees and dirt trails.
%
While the Indoor and Forest categories follow the original M3ED sequence names, we define the Urban categories by grouping diverse urban environments, such as parks, campuses, and roads, from the original labels.
%
Fig.~\ref{fig:HPE_visualize} provides representative RGB examples for each platform and environment category in the HPE scenario, while Tab.~\ref{tab:HPE_sequences_sorted} lists the corresponding sequence-level characteristics, including the train/test split, number of frames, and attributes.
%
Thus, in the HPE scenario, each client corresponds to a single sequence with specific platform and environment attributes.

We construct the BMR scenario by integrating the indoor NYUv2 dataset~\cite{silberman2012indoor} and the outdoor KITTI dataset~\cite{geiger2012we}, using the same preprocessing configuration as DCDepth~\cite{dcdepth}.
To investigate heterogeneity between the indoor and outdoor domains, we assign 2 sequences to each KITTI client and 16 sequences to each NYUv2 client, balancing both the number of clients (15 for NYU, 17 for KITTI) and total data volume across domains.
The characteristics of all sequences are summarized in Tab.~\ref{tab:BMR_sequences_final_v2}.

%

\begin{table*}[htbp]
\centering
\caption{\textbf{Details of the HPE scenario sequences.}
All sequences in the HPE scenario with their train/test split, number of frames, and attributes indicating the robot platform and environment are listed.}
\label{tab:HPE_sequences_sorted}

\resizebox{0.95\textwidth}{!}{
\setlength{\tabcolsep}{8pt}
\begin{tabular}{|c|c|c|c|c|}
\hline
\textbf{Idx} & \textbf{Sequence} & \textbf{Split} & \textbf{Frames} & \textbf{Attributes} \\
\hline

\textbf{1} & car\_forest\_into\_ponds\_short & Train & 1908 & UGV, Forest \\
\textbf{2} & car\_forest\_sand\_1 & Train & 1391 & UGV, Forest \\
\textbf{3} & car\_forest\_tree\_tunnel & Train & 1118 & UGV, Forest \\
\textbf{4} & car\_urban\_day\_city\_hall & Train & 1966 & UGV, Urban-Day \\
\textbf{5} & car\_urban\_day\_horse & Train & 127 & UGV, Urban-Day \\
\textbf{6} & car\_urban\_day\_penno\_big\_loop & Train & 3007 & UGV, Urban-Day \\
\textbf{7} & car\_urban\_day\_penno\_small\_loop & Train & 374 & UGV, Urban-Day \\
\textbf{8} & car\_urban\_day\_rittenhouse & Train & 2347 & UGV, Urban-Day \\
\textbf{9} & car\_urban\_night\_city\_hall & Train & 1886 & UGV, Urban-Night \\
\textbf{10} & car\_urban\_night\_penno\_big\_loop & Train & 3165 & UGV, Urban-Night \\
\textbf{11} & car\_urban\_night\_penno\_small\_loop & Train & 413 & UGV, Urban-Night \\
\textbf{12} & car\_urban\_night\_penno\_small\_loop\_darker & Train & 414 & UGV, Urban-Night \\
\textbf{13} & car\_urban\_night\_rittenhouse & Train & 2651 & UGV, Urban-Night \\
\textbf{14} & falcon\_forest\_into\_forest\_2 & Train & 1179 & UAV, Forest \\
\textbf{15} & falcon\_forest\_into\_forest\_4 & Train & 1612 & UAV, Forest \\
\textbf{16} & falcon\_forest\_road\_1 & Train & 1422 & UAV, Forest \\
\textbf{17} & falcon\_forest\_road\_2 & Train & 1558 & UAV, Forest \\
\textbf{18} & falcon\_forest\_up\_down & Train & 1046 & UAV, Forest \\
\textbf{19} & falcon\_indoor\_flight\_2 & Train & 462 & UAV, Indoor \\
\textbf{20} & falcon\_indoor\_flight\_3 & Train & 452 & UAV, Indoor \\
\textbf{21} & falcon\_outdoor\_day\_fast\_flight\_1 & Train & 670 & UAV, Urban-Day \\
\textbf{22} & falcon\_outdoor\_day\_penno\_cars & Train & 1018 & UAV, Urban-Day \\
\textbf{23} & falcon\_outdoor\_day\_penno\_parking\_1 & Train & 1001 & UAV, Urban-Day \\
\textbf{24} & falcon\_outdoor\_day\_penno\_parking\_2 & Train & 1004 & UAV, Urban-Day \\
\textbf{25} & falcon\_outdoor\_day\_penno\_plaza & Train & 616 & UAV, Urban-Day \\
\textbf{26} & falcon\_outdoor\_day\_penno\_trees & Train & 1269 & UAV, Urban-Day \\
\textbf{27} & falcon\_outdoor\_night\_high\_beams & Train & 431 & UAV, Urban-Night \\
\textbf{28} & falcon\_outdoor\_night\_penno\_parking\_1 & Train & 971 & UAV, Urban-Night \\
\textbf{29} & spot\_forest\_easy\_1 & Train & 658 & Legged, Forest \\
\textbf{30} & spot\_forest\_easy\_2 & Train & 1018 & Legged, Forest \\
\textbf{31} & spot\_forest\_hard & Train & 929 & Legged, Forest \\
\textbf{32} & spot\_forest\_road\_1 & Train & 1382 & Legged, Forest \\
\textbf{33} & spot\_indoor\_obstacles & Train & 770 & Legged, Indoor \\
\textbf{34} & spot\_indoor\_stairs & Train & 786 & Legged, Indoor \\
\textbf{35} & spot\_outdoor\_day\_art\_plaza\_loop & Train & 1306 & Legged, Urban-Day \\
\textbf{36} & spot\_outdoor\_day\_penno\_short\_loop & Train & 1059 & Legged, Urban-Day \\
\textbf{37} & spot\_outdoor\_day\_rocky\_steps & Train & 974 & Legged, Urban-Day \\
\textbf{38} & spot\_outdoor\_day\_skatepark\_1 & Train & 814 & Legged, Urban-Day \\
\textbf{39} & spot\_outdoor\_day\_skatepark\_2 & Train & 591 & Legged, Urban-Day \\
\textbf{40} & spot\_outdoor\_day\_srt\_green\_loop & Train & 496 & Legged, Urban-Day \\
\textbf{41} & spot\_outdoor\_day\_srt\_under\_bridge\_2 & Train & 1709 & Legged, Urban-Day \\

\hline

\textbf{1} & car\_forest\_into\_ponds\_long & Test & 848 & Car, Forest \\
\textbf{2} & car\_urban\_day\_ucity\_small\_loop & Test & 550 & Car, Urban-Day \\
\textbf{3} & car\_urban\_night\_ucity\_small\_loop & Test & 474 & Car, Urban-Night \\
\textbf{4} & falcon\_forest\_road\_forest & Test & 655 & Falcon, Forest \\
\textbf{5} & falcon\_indoor\_flight\_1 & Test & 145 & Falcon, Indoor \\
\textbf{6} & falcon\_outdoor\_day\_fast\_flight\_2 & Test & 644 & Falcon, Urban-Day \\
\textbf{7} & falcon\_outdoor\_night\_penno\_parking\_2 & Test & 277 & Falcon, Urban-Night \\
\textbf{8} & spot\_forest\_road\_3 & Test & 196 & Spot, Forest \\
\textbf{9} & spot\_indoor\_building\_loop & Test & 275 & Spot, Indoor \\
\textbf{10} & spot\_outdoor\_day\_srt\_under\_bridge\_1 & Test & 346 & Spot, Urban-Day \\
\textbf{11} & spot\_outdoor\_night\_penno\_short\_loop & Test & 224 & Spot, Urban-Night \\

\hline
\multicolumn{2}{|c|}{\textbf{Sequences}} & \multicolumn{3}{c|}{\textbf{Total: 52 \quad Train: 41 \quad Test: 11}} \\
\hline
\multicolumn{2}{|c|}{\textbf{Frames}} & \multicolumn{3}{c|}{\textbf{Total: 52,604 \quad Train: 47,970 \quad Test: 4,634}} \\
\hline
\end{tabular}
}
\end{table*}
\begin{table*}[htbp]
\centering
\caption{\textbf{Details of the BMR scenario sequences.} We provide the train/test split, number of frames, and attributes for each sequence.}
\label{tab:BMR_sequences_final_v2}
\resizebox{0.95\textwidth}{!}{%
\setlength{\tabcolsep}{8pt}
\begin{tabular}{|c|c|c|c||c|c|c|c|}
\hline
\multicolumn{4}{|c||}{\textbf{TRAIN SET}} & \multicolumn{4}{c|}{\textbf{TEST SET}} \\
\hline
\textbf{Idx} & \textbf{Sequence} & \textbf{Frames} & \textbf{Attributes} & \textbf{Idx} & \textbf{Sequence} & \textbf{Frames} & \textbf{Attributes} \\
\hline
\textbf{1} & \makecell[c]{2011\_09\_26\_drive\_0001} & 98 & Outdoor & \textbf{1} & \makecell[c]{2011\_09\_26\_drive\_0002} & 25 & Outdoor \\
\textbf{2} & \makecell[c]{2011\_09\_26\_drive\_0005} & 144 & Outdoor & \textbf{2} & \makecell[c]{2011\_09\_26\_drive\_0009} & 25 & Outdoor \\
\textbf{3} & \makecell[c]{2011\_09\_26\_drive\_0011} & 223 & Outdoor & \textbf{3} & \makecell[c]{2011\_09\_26\_drive\_0013} & 25 & Outdoor \\
\textbf{4} & \makecell[c]{2011\_09\_26\_drive\_0014} & 304 & Outdoor & \textbf{4} & \makecell[c]{2011\_09\_26\_drive\_0020} & 25 & Outdoor \\
\textbf{5} & \makecell[c]{2011\_09\_26\_drive\_0015} & 287 & Outdoor & \textbf{5} & \makecell[c]{2011\_09\_26\_drive\_0023} & 25 & Outdoor \\
\textbf{6} & \makecell[c]{2011\_09\_26\_drive\_0017} & 104 & Outdoor & \textbf{6} & \makecell[c]{2011\_09\_26\_drive\_0027} & 25 & Outdoor \\
\textbf{7} & \makecell[c]{2011\_09\_26\_drive\_0018} & 260 & Outdoor & \textbf{7} & \makecell[c]{2011\_09\_26\_drive\_0029} & 25 & Outdoor \\
\textbf{8} & \makecell[c]{2011\_09\_26\_drive\_0019} & 471 & Outdoor & \textbf{8} & \makecell[c]{2011\_09\_26\_drive\_0036} & 25 & Outdoor \\
\textbf{9} & \makecell[c]{2011\_09\_26\_drive\_0022} & 790 & Outdoor & \textbf{9} & \makecell[c]{2011\_09\_26\_drive\_0046} & 25 & Outdoor \\
\textbf{10} & \makecell[c]{2011\_09\_26\_drive\_0028} & 420 & Outdoor & \textbf{10} & \makecell[c]{2011\_09\_26\_drive\_0048} & 22 & Outdoor \\
\textbf{11} & \makecell[c]{2011\_09\_26\_drive\_0032} & 380 & Outdoor & \textbf{11} & \makecell[c]{2011\_09\_26\_drive\_0052} & 25 & Outdoor \\
\textbf{12} & \makecell[c]{2011\_09\_26\_drive\_0035} & 121 & Outdoor & \textbf{12} & \makecell[c]{2011\_09\_26\_drive\_0056} & 25 & Outdoor \\
\textbf{13} & \makecell[c]{2011\_09\_26\_drive\_0039} & 385 & Outdoor & \textbf{13} & \makecell[c]{2011\_09\_26\_drive\_0059} & 25 & Outdoor \\
\textbf{14} & \makecell[c]{2011\_09\_26\_drive\_0051} & 428 & Outdoor & \textbf{14} & \makecell[c]{2011\_09\_26\_drive\_0064} & 25 & Outdoor \\
\textbf{15} & \makecell[c]{2011\_09\_26\_drive\_0057} & 351 & Outdoor & \textbf{15} & \makecell[c]{2011\_09\_26\_drive\_0084} & 25 & Outdoor \\
\textbf{16} & \makecell[c]{2011\_09\_26\_drive\_0060} & 68 & Outdoor & \textbf{16} & \makecell[c]{2011\_09\_26\_drive\_0086} & 25 & Outdoor \\
\textbf{17} & \makecell[c]{2011\_09\_26\_drive\_0061} & 693 & Outdoor & \textbf{17} & \makecell[c]{2011\_09\_26\_drive\_0093} & 25 & Outdoor \\
\textbf{18} & \makecell[c]{2011\_09\_26\_drive\_0070} & 410 & Outdoor & \textbf{18} & \makecell[c]{2011\_09\_26\_drive\_0096} & 25 & Outdoor \\
\textbf{19} & \makecell[c]{2011\_09\_26\_drive\_0079} & 90 & Outdoor & \textbf{19} & \makecell[c]{2011\_09\_26\_drive\_0101} & 25 & Outdoor \\
\textbf{20} & \makecell[c]{2011\_09\_26\_drive\_0087} & 719 & Outdoor & \textbf{20} & \makecell[c]{2011\_09\_26\_drive\_0106} & 25 & Outdoor \\
\textbf{21} & \makecell[c]{2011\_09\_26\_drive\_0091} & 330 & Outdoor & \textbf{21} & \makecell[c]{2011\_09\_26\_drive\_0117} & 25 & Outdoor \\
\textbf{22} & \makecell[c]{2011\_09\_26\_drive\_0095} & 258 & Outdoor & \textbf{22} & \makecell[c]{2011\_09\_28\_drive\_0002} & 25 & Outdoor \\
\textbf{23} & \makecell[c]{2011\_09\_26\_drive\_0104} & 302 & Outdoor & \textbf{23} & \makecell[c]{2011\_09\_29\_drive\_0071} & 25 & Outdoor \\
\textbf{24} & \makecell[c]{2011\_09\_26\_drive\_0113} & 77 & Outdoor & \textbf{24} & \makecell[c]{2011\_09\_30\_drive\_0016} & 25 & Outdoor \\
\textbf{25} & \makecell[c]{2011\_09\_28\_drive\_0001} & 96 & Outdoor & \textbf{25} & \makecell[c]{2011\_09\_30\_drive\_0018} & 25 & Outdoor \\
\textbf{26} & \makecell[c]{2011\_09\_29\_drive\_0004} & 329 & Outdoor & \textbf{26} & \makecell[c]{2011\_09\_30\_drive\_0027} & 25 & Outdoor \\
\textbf{27} & \makecell[c]{2011\_09\_29\_drive\_0026} & 148 & Outdoor & \textbf{27} & \makecell[c]{2011\_10\_03\_drive\_0027} & 25 & Outdoor \\
\textbf{28} & \makecell[c]{2011\_09\_30\_drive\_0020} & 1,094 & Outdoor & \textbf{28} & \makecell[c]{2011\_10\_03\_drive\_0047} & 25 & Outdoor \\
\textbf{29} & \makecell[c]{2011\_09\_30\_drive\_0028} & 5,167 & Outdoor & \textbf{29} & \makecell[c]{bathroom} & 58 & Indoor \\
\textbf{30} & \makecell[c]{2011\_09\_30\_drive\_0033} & 1,584 & Outdoor & \textbf{30} & \makecell[c]{bedroom} & 191 & Indoor \\
\textbf{31} & \makecell[c]{2011\_09\_30\_drive\_0034} & 1,214 & Outdoor & \textbf{31} & \makecell[c]{bookstore} & 11 & Indoor \\
\textbf{32} & \makecell[c]{2011\_10\_03\_drive\_0034} & 4,653 & Outdoor & \textbf{32} & \makecell[c]{classroom} & 23 & Indoor \\
\textbf{33} & \makecell[c]{2011\_10\_03\_drive\_0042} & 1,160 & Outdoor & \textbf{33} & \makecell[c]{computer\_lab} & 3 & Indoor \\
\textbf{34} & \makecell[c]{basement} & 166 & Indoor & \textbf{34} & \makecell[c]{dining\_room} & 55 & Indoor \\
\textbf{35} & \makecell[c]{bathroom} & 1,339 & Indoor & \textbf{35} & \makecell[c]{foyer} & 2 & Indoor \\
\textbf{36} & \makecell[c]{bedroom} & 5,156 & Indoor & \textbf{36} & \makecell[c]{home\_office} & 24 & Indoor \\
\textbf{37} & \makecell[c]{bookstore} & 1,383 & Indoor & \textbf{37} & \makecell[c]{kitchen} & 106 & Indoor \\
\textbf{38} & \makecell[c]{cafe} & 185 & Indoor & \textbf{38} & \makecell[c]{living\_room} & 107 & Indoor \\
\textbf{39} & \makecell[c]{classroom} & 625 & Indoor & \textbf{39} & \makecell[c]{office} & 38 & Indoor \\
\textbf{40} & \makecell[c]{computer} & 36 & Indoor & \textbf{40} & \makecell[c]{office\_kitchen} & 4 & Indoor \\
\textbf{41} & \makecell[c]{conference} & 136 & Indoor & \textbf{41} & \makecell[c]{playroom} & 14 & Indoor \\
\textbf{42} & \makecell[c]{dinette} & 72 & Indoor & \textbf{42} & \makecell[c]{reception\_room} & 5 & Indoor \\
\textbf{43} & \makecell[c]{dining} & 2,592 & Indoor & \textbf{43} & \makecell[c]{study} & 11 & Indoor \\
\textbf{44} & \makecell[c]{exercise} & 87 & Indoor & \textbf{44} & \makecell[c]{study\_room} & 2 & Indoor \\
\textbf{45} & \makecell[c]{foyer} & 35 & Indoor & \textbf{45} & \makecell[c]{} &  &  \\
\textbf{46} & \makecell[c]{furniture} & 973 & Indoor & \textbf{46} & \makecell[c]{} &  &  \\
\textbf{47} & \makecell[c]{home} & 748 & Indoor & \textbf{47} & \makecell[c]{} &  &  \\
\textbf{48} & \makecell[c]{indoor} & 32 & Indoor & \textbf{48} & \makecell[c]{} &  &  \\
\textbf{49} & \makecell[c]{kitchen} & 3,618 & Indoor & \textbf{49} & \makecell[c]{} &  &  \\
\textbf{50} & \makecell[c]{laundry} & 46 & Indoor & \textbf{50} & \makecell[c]{} &  &  \\
\textbf{51} & \makecell[c]{living} & 4,080 & Indoor & \textbf{51} & \makecell[c]{} &  &  \\
\textbf{52} & \makecell[c]{nyu} & 365 & Indoor & \textbf{52} & \makecell[c]{} &  &  \\
\textbf{53} & \makecell[c]{office} & 1,090 & Indoor & \textbf{53} & \makecell[c]{} &  &  \\
\textbf{54} & \makecell[c]{playroom} & 443 & Indoor & \textbf{54} & \makecell[c]{} &  &  \\
\textbf{55} & \makecell[c]{printer} & 56 & Indoor & \textbf{55} & \makecell[c]{} &  &  \\
\textbf{56} & \makecell[c]{reception} & 314 & Indoor & \textbf{56} & \makecell[c]{} &  &  \\
\textbf{57} & \makecell[c]{student} & 181 & Indoor & \textbf{57} & \makecell[c]{} &  &  \\
\textbf{58} & \makecell[c]{study} & 473 & Indoor & \textbf{58} & \makecell[c]{} &  &  \\
\hline
\multicolumn{4}{|c||}{\textbf{Sequences}} & \multicolumn{4}{c|}{\textbf{Total: 102 \quad Train: 58 \quad Test: 44}} \\
\hline
\multicolumn{4}{|c||}{\textbf{Frames}} & \multicolumn{4}{c|}{\textbf{Total: 48,740 \quad Train: 47,389 \quad Test: 1,351}} \\
\hline
\end{tabular}%
}
\end{table*}
%
\begin{table*}[p]
\centering
\caption{
\textbf{Clustering results of \textsc{FeDepth} for NeWCRFs~\cite{newcrfs} on the HPE scenario.}
Sequences assigned to each cluster during training and inference. 
Attributes denote the platform and environment.
}
\label{tab:clustering_optimized}

\resizebox{!}{0.45\textheight}{
\setlength{\tabcolsep}{12pt}
\begin{tabular}{|c|c|c|c|}
\hline
\textbf{Cluster} & \textbf{Split} & \textbf{Sequence} & \textbf{Attributes} \\
\hline

\multirow{5}{*}{0}
& \multirow{4}{*}{Train}
& car\_forest\_into\_ponds\_short & \multirow{4}{*}{UGV, Legged / Forest}\\
& & car\_forest\_sand\_1 & \\
& & car\_forest\_tree\_tunnel & \\
& & spot\_forest\_hard & \\
\cline{2-4}
& Test & car\_forest\_into\_ponds\_long & UGV / Forest \\
\hline

\multirow{4}{*}{1}
& \multirow{3}{*}{Train}
& car\_urban\_day\_city\_hall & \multirow{3}{*}{UGV / Urban-Day}\\
& & car\_urban\_day\_horse & \\
& & car\_urban\_day\_rittenhouse & \\
\cline{2-4}
& Test & car\_urban\_day\_ucity\_small\_loop & UGV / Urban-Day \\
\hline

\multirow{5}{*}{2}
& \multirow{4}{*}{Train}
& car\_urban\_day\_penno\_big\_loop & \multirow{4}{*}{UGV, UAV / Urban-Day}\\
& & car\_urban\_day\_penno\_small\_loop & \\
& & falcon\_outdoor\_day\_fast\_flight\_1 & \\
& & falcon\_outdoor\_day\_penno\_plaza & \\
\cline{2-4}
& Test & - & - \\
\hline

\multirow{3}{*}{3}
& \multirow{2}{*}{Train}
& car\_urban\_night\_city\_hall & \multirow{2}{*}{UGV / Urban-Night}\\
& & car\_urban\_night\_rittenhouse & \\
\cline{2-4}
& Test & car\_urban\_night\_ucity\_small\_loop & UGV / Urban-Night \\
\hline

\multirow{5}{*}{4}
& \multirow{3}{*}{Train}
& car\_urban\_night\_penno\_big\_loop & \multirow{3}{*}{UGV, UAV / Urban-Night}\\
& & falcon\_outdoor\_night\_high\_beams & \\
& & falcon\_outdoor\_night\_penno\_parking\_1 & \\
\cline{2-4}
& \multirow{2}{*}{Test}
& falcon\_outdoor\_night\_penno\_parking\_2 & \multirow{2}{*}{UAV, Legged / Urban-Night}\\
& & spot\_outdoor\_night\_penno\_short\_loop & \\
\hline

\multirow{3}{*}{5}
& \multirow{2}{*}{Train}
& car\_urban\_night\_penno\_small\_loop & \multirow{2}{*}{UGV / Urban-Night}\\
& & car\_urban\_night\_penno\_small\_loop\_darker & \\
\cline{2-4}
& Test & spot\_outdoor\_night\_penno\_short\_loop & Legged / Urban-Night \\
\hline

\multirow{2}{*}{6}
& Train & falcon\_forest\_into\_forest\_\{2,4\} & UAV / Forest \\
\cline{2-4}
& Test & - & - \\
\hline

\multirow{4}{*}{7}
& \multirow{3}{*}{Train}
& falcon\_forest\_road\_\{1,2\} & \multirow{3}{*}{UAV, Legged / Forest}\\
& & falcon\_forest\_up\_down & \\
& & spot\_forest\_hard & \\
\cline{2-4}
& Test & falcon\_forest\_road\_forest & UAV / Forest \\
\hline

\multirow{4}{*}{8}
& \multirow{3}{*}{Train}
& falcon\_forest\_road\_\{1,2\} & \multirow{3}{*}{UAV, Legged / Forest}\\
& & falcon\_forest\_up\_down & \\
& & spot\_forest\_easy\_2 & \\
\cline{2-4}
& Test & falcon\_forest\_road\_forest & UAV / Forest \\
\hline

\multirow{3}{*}{9}
& Train & falcon\_indoor\_flight\_\{2,3\} & UAV / Indoor \\
\cline{2-4}
& Test & falcon\_indoor\_flight\_1 & UAV / Indoor \\
\hline

\multirow{3}{*}{10}
& \multirow{2}{*}{Train}
& falcon\_outdoor\_day\_penno\_cars & \multirow{2}{*}{UAV / Urban-Day}\\
& & falcon\_outdoor\_day\_penno\_trees & \\
\cline{2-4}
& Test & falcon\_outdoor\_day\_fast\_flight\_2 & UAV / Urban-Day \\
\hline

\multirow{2}{*}{11}
& Train & falcon\_outdoor\_day\_penno\_parking\_\{1,2\} & UAV / Urban-Day \\
\cline{2-4}
& Test & - & - \\
\hline

\multirow{4}{*}{12}
& \multirow{3}{*}{Train}
& falcon\_forest\_up\_down & \multirow{3}{*}{UAV, Legged / Forest}\\
& & spot\_forest\_easy\_\{1,2\} & \\
& & spot\_forest\_hard & \\
\cline{2-4}
& Test & falcon\_forest\_road\_forest & UAV / Forest \\
\hline

\multirow{7}{*}{13}
& \multirow{6}{*}{Train}
& car\_forest\_into\_ponds\_short & \multirow{6}{*}{UGV, UAV, Legged / Forest}\\
& & car\_forest\_tree\_tunnel & \\
& & falcon\_forest\_road\_2 & \\
& & spot\_forest\_easy\_\{1,2\} & \\
& & spot\_forest\_hard & \\
& & spot\_forest\_road\_1 & \\
\cline{2-4}
& \multirow{2}{*}{Test}
& falcon\_forest\_road\_forest & \multirow{2}{*}{UAV, Legged / Forest}\\
& & spot\_forest\_road\_3 & \\
\hline

\multirow{3}{*}{14}
& \multirow{2}{*}{Train}
& spot\_indoor\_obstacles & \multirow{2}{*}{Legged / Indoor}\\
& & spot\_indoor\_stairs & \\
\cline{2-4}
& Test & spot\_indoor\_building\_loop & Legged / Indoor \\
\hline

\multirow{3}{*}{15}
& \multirow{2}{*}{Train}
& spot\_outdoor\_day\_art\_plaza\_loop & \multirow{2}{*}{Legged / Urban-Day}\\
& & spot\_outdoor\_day\_rocky\_steps & \\
\cline{2-4}
& Test & - & - \\
\hline

\multirow{2}{*}{16}
& Train & spot\_outdoor\_day\_skatepark\_\{1,2\} & Legged / Urban-Day \\
\cline{2-4}
& Test & - & - \\
\hline

\multirow{4}{*}{17}
& \multirow{3}{*}{Train}
& spot\_outdoor\_day\_penno\_short\_loop & \multirow{3}{*}{Legged / Urban-Day}\\
& & spot\_outdoor\_day\_srt\_green\_loop & \\
& & spot\_outdoor\_day\_srt\_under\_bridge\_2 & \\
\cline{2-4}
& Test & spot\_outdoor\_day\_srt\_under\_bridge\_1 & Legged / Urban-Day \\
\hline

\end{tabular}}
\end{table*}

%

\end{document}